\documentclass[10pt,twocolumn,letterpaper]{article}

\usepackage[pagenumbers]{cvpr} 
\usepackage{algorithm}
\usepackage{algpseudocode}
\usepackage{multirow}

\definecolor{cvprblue}{rgb}{0.21,0.49,0.74}
\usepackage[pagebackref,breaklinks,colorlinks,allcolors=cvprblue]{hyperref}

\usepackage{amsmath,amsfonts,bm}

\def\eqref#1{equation~\ref{#1}}

\def\ve{{\bm{e}}}

\def\vg{{\bm{g}}}

\def\vt{{\bm{t}}}
\def\vu{{\bm{u}}}

\def\vx{{\bm{x}}}
\def\vy{{\bm{y}}}
\def\vz{{\bm{z}}}

\def\mA{{\bm{A}}}

\def\mI{{\bm{I}}}

\DeclareMathAlphabet{\mathsfit}{\encodingdefault}{\sfdefault}{m}{sl}
\SetMathAlphabet{\mathsfit}{bold}{\encodingdefault}{\sfdefault}{bx}{n}

\newcommand{\dec}{\ensuremath{Dec}}
\newcommand{\enc}{\ensuremath{Enc}}
\newcommand*\samethanks[1][\value{footnote}]{\footnotemark[#1]}

\title{ZoomLDM: Latent Diffusion Model for multi-scale image generation}

\author{
    Srikar Yellapragada\thanks{Equal contribution. Correspondence to \href{mailto:srikary@cs.stonybrook.edu}{srikary@cs.stonybrook.edu}} \quad Alexandros Graikos\samethanks[1] \quad Kostas Triaridis \\
    Prateek Prasanna \quad Rajarsi R. Gupta \quad Joel Saltz \quad Dimitris Samaras \\ \\
    Stony Brook University
}

\begin{document}
\maketitle

\begin{abstract}
Diffusion models have revolutionized image generation, yet several challenges restrict their application to large-image domains, such as digital pathology and satellite imagery. Given that it is infeasible to directly train a model on 'whole' images from domains with potential gigapixel sizes, diffusion-based generative methods have focused on synthesizing small, fixed-size patches extracted from these images. However, generating small patches has limited applicability since patch-based models fail to capture the global structures and wider context of large images, which can be crucial for synthesizing (semantically) accurate samples. To overcome this limitation, we present ZoomLDM, a diffusion model tailored for generating images across multiple scales. Central to our approach is a novel magnification-aware conditioning mechanism that utilizes self-supervised learning (SSL) embeddings and allows the diffusion model to synthesize images at different 'zoom' levels, i.e., fixed-size patches extracted from large images at varying scales. ZoomLDM synthesizes coherent histopathology images that remain contextually accurate and detailed at different zoom levels, achieving state-of-the-art image generation quality across all scales and excelling in the data-scarce setting of generating thumbnails of entire large images. The multi-scale nature of ZoomLDM unlocks additional capabilities in large image generation, enabling computationally tractable and globally coherent image synthesis up to $4096 \times 4096$ pixels and $4\times$ super-resolution. Additionally, multi-scale features extracted from ZoomLDM are highly effective in multiple instance learning experiments. \footnote{Code is available at \href{https://github.com/cvlab-stonybrook/ZoomLDM}{github.com/cvlab-stonybrook/ZoomLDM}.}
\end{abstract}

\section{Introduction}
\label{sec:intro}

\begin{figure}[ht]
    \includegraphics[width=\columnwidth]{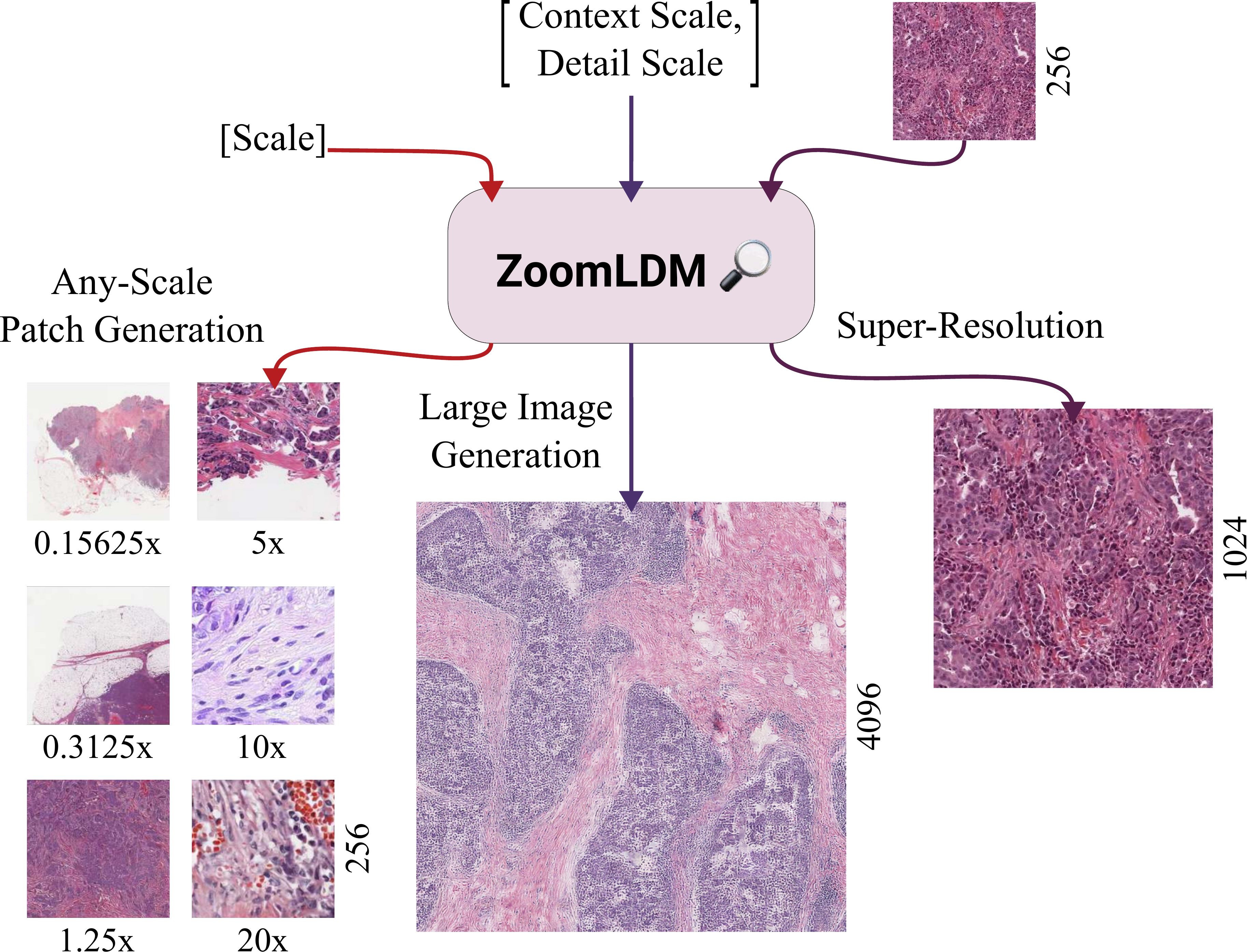}
    \centering
    \caption{ZoomLDM can generate synthetic image patches at multiple scales (left). It can generate large images that preserve spatial context (center) and perform super-resolution (right), without any additional training. Large images from prior work \cite{le2024inftybrushcontrollablelargeimage, graikos2024learned} suffer from blurriness and lack of global context.}
    \label{fig:teaser}
    \vspace{-0.5cm}
\end{figure}

Diffusion models have achieved remarkable success in photorealistic image synthesis \cite{betker2023improving}, benefiting from the availability of vast multi-modal datasets \cite{schuhmann2022laion,changpinyo2021conceptual} and sophisticated conditioning techniques \cite{ho2022classifier,podell2024sdxl}. Latent Diffusion models (LDMs) \cite{rombach2022high} have further advanced high-resolution image generation by introducing a two-step process that first compresses the images with a learned encoder and then trains the generative diffusion model in that encoder's latent space. In the natural image domain, LDMs like Stable Diffusion XL \cite{podell2024sdxl}, which generates $1024 \times 1024$ images, have made high-resolution generation fast and cheap. Although such models demonstrate the potential of further scaling image diffusion to larger sizes, large-image domains such as digital histopathology and satellite imagery are beyond their feasible scope as images there are typically in the gigapixel scale (e.g. $32,000 \times 32,000$ pixels).

Apart from scale, large-image domains also lack paired image-annotation data with sufficient detail, which has been key to the success of text-to-image diffusion models. Without access to a conditioning signal during training and inference, the performance of diffusion models degrades significantly \cite{nichol2022glide}. At the same time, obtaining annotations for large images can be complex as it is both a laborious process for specialized fields, such as medical images, and often ambiguous since annotators can describe different features at different scales. A satellite image text caption corresponding to \emph{`water'}, when viewed from up close, can turn into both the \emph{`a lake'} and \emph{`a river'} when viewed from further away, making it necessary to annotate at both levels.

Previous works have tried to address the issues of large image sizes and conditioning but are limited in applicability. \citet{Harb_2024_WACV} introduced a pixel-level diffusion model that can accommodate multiple scales (named magnifications) in medical images but lacked conditioning - a crucial element for achieving better image quality and enabling downstream tasks \cite{nichol2021improved,dhariwal2021diffusion,Yellapragada_2024_WACV}. \citet{graikos2024learned} utilized embeddings from self-supervised learning (SSL) models to mitigate the need for costly annotations in large-image domains, but only trained a model to generate small patches. Recognizing that none of these methods can tackle the important problem of \emph{controllable high-quality large-image synthesis}, we propose a unified solution, ZoomLDM.

To address large image sizes, we propose training a scale-conditioned diffusion model that learns to generate images at different `zoom' levels, which correspond to magnifications in histopathology images (Fig.~\ref{fig:teaser} (a)). By conditioning the model on the scale, we control the level of detail contained within each generated pixel. To control generation, we also incorporate a conditioning signal from a self-supervised learning (SSL) encoder. While SSL encoders are great at producing meaningful representations for images, using them in this multi-scale setting is nontrivial as they are usually trained to extract information from patches at a single scale. To share information across scales, we introduce the idea of a cross-magnification latent space; a shared latent space where the embeddings of all scales lie. We implement this with
a trainable \emph{summarizer} module that processes the array of SSL embeddings that describe an image, projecting them to the shared latent space that captures dependencies across all magnifications.

We train ZoomLDM on multi-scale histopathology using SSL embeddings from state-of-the-art image encoders as guidance. We find that sharing model weights across all scales significantly boosts the generation quality for scales where data is limited. To eliminate our model's reliance on SSL embeddings when sampling new images, we also train a Conditioning Diffusion Model (CDM) that generates conditions given a scale. This combined approach enables us to synthesize novel high-quality images at all scales.

With a multi-scale model, we hypothesize that jointly sampling images across scales would be beneficial for creating coherent images at multiple scales. However, this is challenging because each scale requires its own level of detail, and these details must be aligned across scales. To that end, we propose a novel joint multi-scale sampling approach that exploits ZoomLDM's multi-scale nature. Our cross-magnification latent space provides the necessary detail across scales, enabling large image generation and super-resolution without additional training. This approach effectively constructs a coherent image pyramid, making super-resolution and high-quality large image generation feasible.  Our method surpasses previous approaches \cite{le2024inftybrushcontrollablelargeimage, graikos2024learned}, which struggled in generating either local details or global structure, and presents the first practical $4096 \times 4096$ image generation paradigm in histopathology (see supplementary for a comprehensive evaluation).

Finally, we probe ZoomLDM to show that features extracted from our model are highly expressive and suitable for multiple instance learning (MIL) tasks in digital histopathology. Prior work \cite{li2021dual, chen2022scaling} has demonstrated the effectiveness of multi-scale features for MIL, but these methods required training separate encoders for each scale. In contrast, ZoomLDM offers an efficient solution by enabling seamless multi-scale feature extraction using a single model. 
We condition ZoomLDM with UNI\cite{chen2024uni}, a SoTA SSL model, and extract intermediate features from the denoiser at multiple magnifications for MIL. As expected, fusing ZoomLDM features from multiple scales outperforms using SoTA encoders in our MIL experiments, displaying the efficacy of its multi-scale representations. Surprisingly, our features from just the $20 \times$ magnification alone surpass UNI features. We hypothesize that by learning to generate at multiple scales, ZoomLDM has learned to produce more informative features.

Our contributions are the following:
\begin{itemize}
    \item We present \textbf{ZoomLDM}, the first multi-scale conditional latent diffusion model that generates images at multiple scales, achieving state-of-the-art synthetic image quality.
    \item We introduce a cross-magnification latent space, implemented with a trainable summarizer module, which provides conditioning across scales, allowing ZoomLDM to capture dependencies across magnifications. 
    \item We propose a novel joint multi-scale sampling approach for generating large images that retain both global context and local fidelity, making us the first to efficiently synthesize good quality histopathology image samples of up to $4096 \times 4096$ pixels. 
    \item We probe the learned multi-scale representations of ZoomLDM and demonstrate their usefulness by surpassing SoTA encoders on multiple instance learning tasks.
\end{itemize}

\begin{figure*}[ht]
    \includegraphics[width=\textwidth]{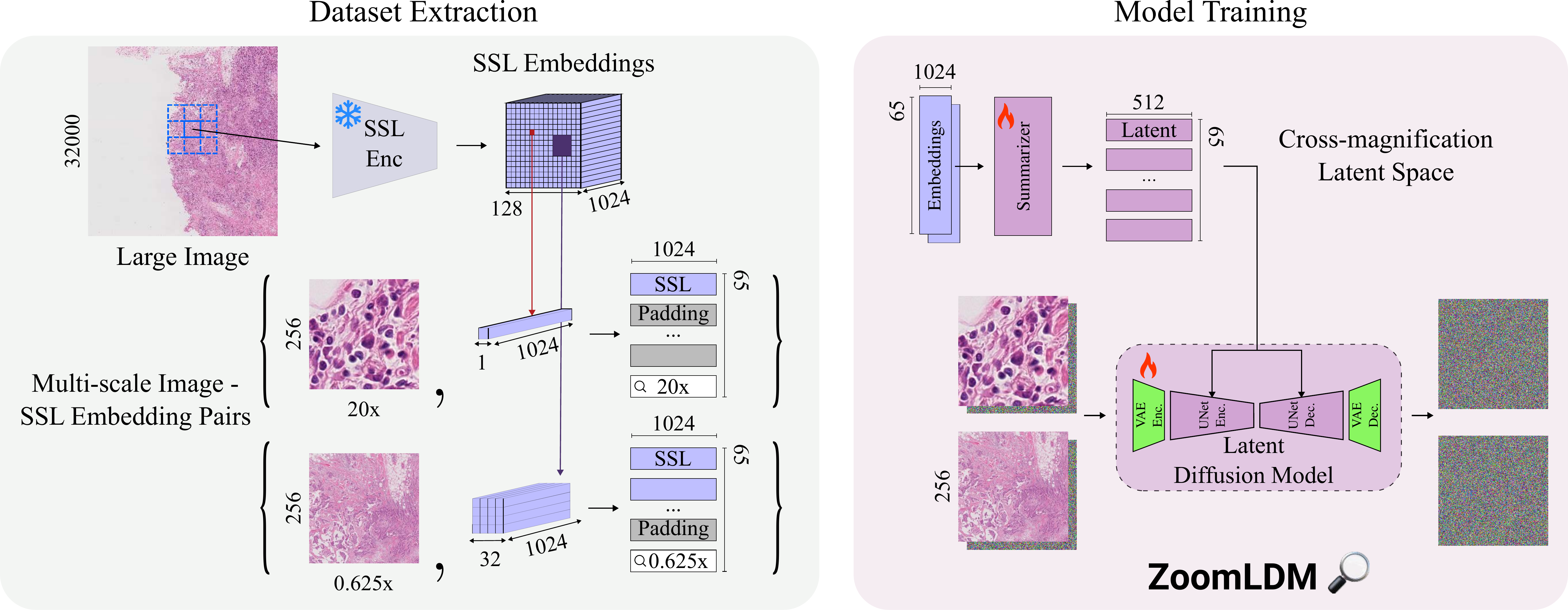}
    \centering
    \caption{Overview of our approach. Left: We extract $256 \times 256$ patches from large images at the initial scale ($20\times$ for pathology) and generate SSL embedding matrices using pretrained encoders. The large image is then progressively downsampled by a factor of 2, with patches at each scale paired with the SSL embeddings of all overlapping initial-scale patches. Right: The SSL embeddings and magnification level are fed to the Summarizer, which projects them into the cross-magnification Latent space. The diffusion model is trained to generate $256 \times 256$ patches conditioned on the Summarizer's output.}
    \label{fig:method}
\end{figure*}

\section{Related Work}
\label{sec:related}
\textbf{Diffusion models:} Since their initial introduction to image generation in \citet{ho2020denoising}, diffusion models have become the dominant generative models for images. Several works have been pivotal; notably class conditioning~\cite{nichol2021improved} which highlighted the importance of guidance during training and sampling and its extensions with classifier~\cite{dhariwal2021diffusion} and classifier-free guidance~\cite{ho2022classifier}. Latent Diffusion Models (LDMs)~\cite{rombach2022high} proposed a training the diffusion model in a Variational Autoencoder (VAE) latent space, compressing the input images by a factor of up to $\times 8$ and enabling high-resolution and computationally practical image generation. Denoising Diffusion Implicit Models (DDIM)~\cite{song2020denoising} accelerated the sampling process further, making diffusion models the preferred alternative over all previous generative model approaches (GANs, Normalizing Flows).

\textbf{Diffusion Models in Large-Image Domains:} Despite advances in the domain of natural images, training generative models directly at the gigapixel resolution of large image domains remains infeasible. Proposed alternatives generate images in a coarse-to-fine process by chaining models in a cascading manner ~\cite{podell2023sdxl,saharia2022image}. This has led to synthesizing images of up to $1024 \times 1024$ resolution at the cost of increased parameter count and slower inference speed. Recently, PixArt-$\Sigma$ \cite{chen2024pixartsigma} introduced an efficient transformer architecture that enables image generation of up to $4k$ using a weak-to-strong training strategy.

In the context of histopathology, previous works have focused on training fixed-size, patch diffusion models \cite{moghadam2023morphology,xu2023vit,muller2023multimodal,Yellapragada_2024_WACV}, with similar approaches in satellite data \cite{espinosa2023generate, sebaq2023rsdiff}. Patch models were used to extrapolate to large images in \cite{aversa2023diffinfinite}, where a pre-generated segmentation mask guides the patch model over the large image, and \cite{graikos2024learned} where a patch model is conditioned on SSL embeddings that smoothly vary across the large image, synthesizing appearance locally. Both methods fail to understand global structures and rely on external sources of information for guidance.

More closely related to our work, \cite{Harb_2024_WACV} trains a pathology diffusion model conditioned on image scales. However, limited evaluations and the absence of a conditioning mechanism restrict its applicability. A different approach by \citet{le2024inftybrushcontrollablelargeimage} utilized an infinite-dimensional diffusion model that is resolution-free, meaning that it can be trained on arbitrarily large images. Their model can be scaled for up to $4096 \times 4096$ generation, but the final results are usually blurry and lack details.

\section{Method}
\subsection{Unified Multi-Scale Training}
We train ZoomLDM to generate fixed-size $256 \times 256$ patches extracted at different scales of large images. To guide generation, we introduce a novel conditioning mechanism allowing the model to learn multi-scale dependencies. Figure~\ref{fig:method} provides an overview of our multi-scale training.

We begin by extracting $256 \times 256$ image patches from a large image at full resolution. Since there are no descriptive patch-level annotations in large-image domains, we resort to pre-trained SSL encoders to provide detailed descriptors in place of human labels, as in \cite{graikos2024learned}. The SSL encoders in these domains are usually trained on patches from these large images -- for histopathology, we utilize UNI \cite{chen2022scaling}, an image encoder trained on $224\times 224$ px $20\times$ patches. After extracting patches $\mI^1$ at the initial scale (=1) and SSL embeddings $\ve$, we end up with a dataset of $\{\mI^1_i,\ve_i\}$ pairs.

We downsample the large image by a factor of 2 and repeat the patch extraction process, getting a new set of patches at the next zoom level. But, as previously mentioned, we cannot directly use the SSL encoder on images from different scales -- e.g., UNI is only trained on $20\times$ images. Therefore, for scales above the first, we utilize the embeddings corresponding to the region contained within the context of the current-scale patch as conditioning. This means that we pair $\mI^2$ patches with the embeddings of all the $\mI^1$ images that they contain, giving us a dataset of 
$\{\mI^2_i\bigl( \begin{smallmatrix} 
  \ve_1 & \ve_2\\
  \ve_3 & \ve_4 
\end{smallmatrix} \bigr)_i\}$
pairs.

By repeating this process, we construct a dataset of (image, embeddings) pairs for all scales, which we want to utilize as our training data for a latent diffusion model. The issue is that the number of SSL embeddings for an image size increases exponentially as we increase scale. This leads to significant computational overhead, primarily due to the quadratic complexity of cross-attention mechanisms used to condition diffusion models. Additionally, conditioning the generation of $256 \times 256$ images with a massive number of embeddings is redundant, given that if we have a total of 8 scales then we will be using a $128 \times 128 \times D$ condition to generate a single $256 \times 256 \times 3$ patch.

To address this issue, we introduce the idea of a learned cross-magnification latent space, shared across embeddings of all scales. To implement this, we train a ``\emph{Summarizer}" transformer, jointly with the diffusion denoiser, that processes the SSL embeddings extracted alongside every image. The information contained in the embeddings is ``summarized" in conjunction with an embedding of the image scale, extracting the essential information needed by the LDM to synthesize patches accurately.

The variable number of tokens (embeddings) in the summarizer input is transformed into a fixed-sized set of conditioning tokens. We utilize padding and pooling to provide a fixed-size output with which we train the LDM. The magnification embedding added to the input makes the summarizer scale-aware, allowing it to adapt to the appropriate level of detail required at different scales. The output of the Summarizer then serves as conditioning input for the LDM, enabling the model to generate high-quality patches with scale-adaptive conditioning.

\paragraph{Conditioning Diffusion Model.}
Our image synthesis pipeline requires a set of SSL embeddings and the desired magnification level, which involves extracting the conditioning information from reference real large-images. This becomes impractical when direct access to training data is unavailable. To address this, we train a second diffusion model, the Conditioning Diffusion Model (CDM), which learns to sample from the distribution of the learned cross-magnification latent space after training the LDM.

Rather than training a diffusion model to model the distribution of the SSL embeddings, which is as complex as learning the distribution of images, we learn the output of the Summarizer, as it captures the most relevant information for synthesizing an image at a given magnification. This approach allows the CDM to model a more refined, task-specific latent space. By also conditioning the CDM on scale, we enable magnification-aware novel image synthesis, which we show can generate high-quality, non-memorized images at the highest scale, even if the amount of data is incredibly scarce (2500 images at $0.15625\times$ magnification).

\subsection{Joint Multi-Scale Sampling}
\label{subsec:joint_sampling}
One of the biggest challenges in large-image domains is synthesizing images that contain local details and exhibit global consistency. Due to their immense sizes, we cannot directly train a model on the full gigapixel images, and training on individual scales will either lead to loss of detail or contextually incoherent results. 

We propose a multi-scale training pipeline intrinsically motivated by the need to sample images from multiple scales jointly.
By drawing samples jointly, we can balance the computational demands of generating large images by separating the global context generation, which is offset by synthesizing an image at a coarser scale, and the synthesis of fine local details, which is done at the lowest level.

We develop a joint multi-scale sampling approach that builds upon ZoomLDM's multi-scale nature and enables us to generate large images of up to $4096 \times 4096$ pixels. The key to our approach is providing 'self-guidance' to the model by guiding the generation of the lowest scales using the so-far-generated global context. To implement this guidance we build upon a recent diffusion inference algorithm \cite{graikos2024fast}, which enables fast conditional inference.

\noindent \textbf{Inference Algorithm} An image at scale $s+1$ corresponds to four images at the previous scale $s$ since, during training, we downsample the large images by a factor of 2 at every scale. We want to jointly generate the four patches at the smaller scale $\vx^s_i,\ i=1,\dots,3$ and the single image at the next level $\vx^{s+1}$. The relationship between these images is known; we can recover $\vx^{s+1}$ by multiplying with a linear downsampling operator $\mA$:
\begin{equation}
  \vx^{s+1} = \mA 
  \begin{pmatrix} 
    \vx^s_1 & \vx^s_2\\
    \vx^s_3 & \vx^s_4 
  \end{pmatrix}.
\label{eq:scale_downsampling}
\end{equation}
We use the above matrix notation to denote the spatial arrangement of images. The algorithm proposed in \cite{graikos2024fast} introduces a method to sample an image from a diffusion model given a linear constraint. Given that our multi-scale images are related by a linear constraint, we use a modified version of this algorithm to perform joint sampling across magnifications. We first provide a brief overview and then present the modifications necessary for joint multi-scale sampling. 

Since we use an LDM, we perform the denoising in the VAE latent space and require the $\dec$ and $\enc$ networks to map from latents $\vz$ to images $\vx$ and back. The algorithm requires a linear operator $\mA$ (and its transpose $\mA^T$) and a pixel-space measurement $\vy$ that we want our final sample $\vz_0$ to match, minimizing $C = || \mA\dec(\vz_0) - \vy ||_2^2$. In every step $t$ of the diffusion process, the current noisy latent $\vz_t$ is used to estimate the final 'clean' latent $\hat{\vz}_0(\vz_t)$, by applying the denoiser model $\bm{\epsilon}_{\theta}(\vz_t)$ and Tweedie's formula \cite{efron2011tweedie}. In the typical DDIM \cite{song2020denoising} sampling process, the next diffusion step is predicted as
\begin{equation}
    \vz_{t-1} = \sqrt{\bar{\alpha}_t} \hat{\vz_0}(\vz_t) \sqrt{1-\bar{\alpha_t}} \bm{\epsilon}_{\theta}(\vz_t) + \tilde{\beta_t}\bm{\epsilon}_t.
    \label{eq:ddim}
\end{equation}
The algorithm of \cite{graikos2024fast} proposes minimizing the $C(\vz_t) =  || \mA\dec(\hat{\vz_0}(\vz_t)) - \vy ||_2^2$ w.r.t. $\vz_t$ at every timestep $t$ before performing the DDIM step. To do that it first computes an error direction as
\begin{equation}
    \ve = \nabla \hat{\vz_0} || \mA \dec(\hat{\vz}_0(\vz_t)) - \vy ||_2^2.
    \label{eq:error_dir}
\end{equation}
This error direction and the current noisy sample $\vz_t$ are used to compute the gradient  $\vg = \nabla_{\vz_t} C(\vz_t) =  \nabla_{\vz_t} || \mA\hat{\vz_0}(\vz_t) - \vy ||_2^2$ using a finite difference approximation and the current noisy sample $\vz_t$ is updated:
\begin{gather}
    \vg \approx \left[ \hat{\vz}_0(\vz_t + \delta \ve) - \hat{\vz}_0(\vz_t) \right] / \delta,\label{eq:gradient_approx} \\
    \vz_t \leftarrow \vz_t + \lambda \vg. \label{eq:gradient_update}
\end{gather}

\noindent \textbf{Efficient Joint Sampling} We make two significant modifications to this algorithm to perform the joint multi-scale sampling. First, since we do not have access to a real measurement $\vy$, which corresponds to the higher scale image $\vx^{s+1}$, we use the estimate of the image $\dec(\hat{\vz}^{s+1})$ to guide the generation of $z^s$. Second, we propose a more efficient way of computing error direction (Eq.~\ref{eq:error_dir}), which does not require memory and time-intensive backpropagations. To jointly sample images from scales $s+4$ and $s$ we need to generate $16\times16+1$ total images, which would be infeasible with the previous error computation.

To avoid the backpropagation during (Eq.~\ref{eq:error_dir}) we propose computing a numerical approximation of $\ve$. Similar to Eq.~\ref{eq:gradient_update} we utilize finite differences and compute
\begin{equation}
    \ve \approx \left[ \enc(\dec(\hat{\vz_0}) + \zeta\ve_{img}) - \enc(\dec(\hat{\vz_0})) \right] / \zeta
    \label{eq:vae_error}
\end{equation}
where $\ve_{img} = \mA^T(\mA\dec(\hat{\vx_0}(\vx_t)) - \vy)$. This eliminates the need to backpropagate through the decoder without significantly sacrificing the quality of the images generated. We provide a detailed background of the conditional inference algorithm and how our approximation reduces computation in the supplementary material.

\section{Experiments}

In this section, we showcase the experiments conducted to validate the effectiveness of our method. We train the unified latent diffusion model, ZoomLDM, on patches from eight different magnifications in histopathology. We evaluate the quality of synthetic samples using both real and CDM-sampled conditions. Further, we exploit the multi-scale nature of ZoomLDM to demonstrate its strength in generating good quality high-resolution images across scales, and its utility in super-resolution (SR) and multiple instance learning (MIL) tasks.

\subsection{Setup}

\subsubsection{Implementation details}

We train the LDMs on 3 NVIDIA H100 GPUs, with a batch size 200 per GPU. We use the training code and checkpoints provided by \cite{rombach2022high}. Our LDM configuration consists of a VQ-f4 autoencoder and a U-Net model pre-trained on ImageNet. We set the learning rate at $10^{-4}$ with a warmup of 10,000 steps. The Summarizer is implemented as a 12-layer Transformer, modeled after ViT-Base. For the CDM, we train a Diffusion Transformer \cite{peebles2023scalable} on the outputs of the summarizer. We utilize DDIM sampling \cite{song2020denoising} with 50 steps for both models and apply classifier-free guidance \cite{ho2022classifier} sampling with a scale of 2.0 to create synthetic images. See supplemental for more details on the Summarizer and CDM.

\subsubsection{Dataset}
We select 1,136 whole slide images (WSI) from TCGA-BRCA \cite{cancer2013cancer}. Using the code from DSMIL\cite{li2021dual}, we extract $256 \times 256$ patches at eight different magnifications: $20\times$, $10\times$, $5\times$, $2.5\times$, $1.25\times$, $0.625\times$, $0.3125\times$, and $0.15625\times$. Each patch is paired with its corresponding base resolution ($20\times$) region—for instance, a $256 \times 256$ pixel patch at 5x magnification is paired with a $1024 \times 1024$ pixel region at $20\times$. We then process the $20\times$ regions through the UNI encoder \cite{chen2023general} to produce an embedding matrix for each patch.

The dimensions of this embedding matrix vary based on the patch's magnification level. For example, a $5\times$ patch corresponding to a $20\times$ region of size $1024 \times 1024$   results in an embedding matrix of dimensions $4 \times 4 \times 1024$. As discussed previously, to avoid redundancy in large embedding matrices, we average pool embeddings larger than $8 \times 8$ to $8 \times 8$ (magnifications 1.25 $\times$ and lower). 

In the supplementary, we also provide results for training ZoomLDM on satellite images. We use a similar training setting, replacing the WSIs from pathology with NAIP \cite{naip} tiles and the SSL encoder with DINO-v2 \cite{oquab2023dinov2}, showing the wider applicability of the proposed model.

\begin{table*}[ht]
\centering
\caption{FID of patches generated from ZoomLDM across different magnifications, compared with single magnification models. ZoomLDM achieved better FID scores than SoTA, with particularly significant improvements at lower scales. }
\begin{tabular}{|c|c|c|c|c|c|c|c|c|}
\hline
Magnification       & $20\times$     & $10\times$    & $5\times$    & $2.5\times$   & $1.25\times$  & $0.625\times$ & $0.3125\times$ & $0.15625\times$ \\ \hline
\# Training patches & 12 Mil & 3 Mil & 750k & 186k  & 57k   & 20k   & 7k     & 2.5k    \\ \hline

ZoomLDM    & \textbf{6.77}  & \textbf{7.60}  & \textbf{7.98} &  \textbf{10.73}     &   \textbf{8.74}    & \textbf{7.99} & \textbf{8.34} &\textbf{13.42} \\ 

  SoTA    & 6.98 \cite{graikos2024learned}   & 7.64 \cite{Yellapragada_2024_WACV}  & 9.74 \cite{graikos2024learned} &  20.45     &   39.72    & 58.98 & 66.28  & 106.14  \\ \hline
  \hline

  CDM  & 9.04 & 10.05 & 14.36 & 19.68 & 14.06 & 13.46  & 14.40  &  26.09  \\
\hline
\end{tabular}

\label{tab:fid}
\end{table*}

\begin{figure*}[ht]
    \includegraphics[width=\linewidth]{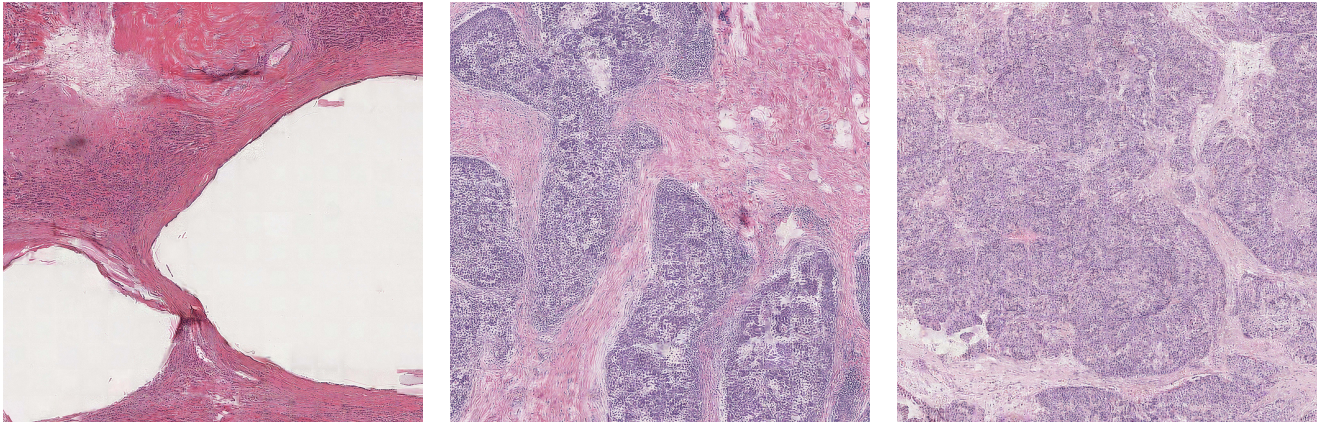}
    \centering
    \caption{Large Images ($4096 \times 4096$) generated from ZoomLDM. Our large image generation framework is the first to generate 4k pathology images with \textbf{local details} and \textbf{global consistency}, all within reasonable inference time. We provide more 4k examples and comparisons in the supplementary.}
    \label{fig:large-image}
    \vspace{-0.5cm}
\end{figure*}

\begin{figure*}[ht]
    \includegraphics[width=\textwidth]{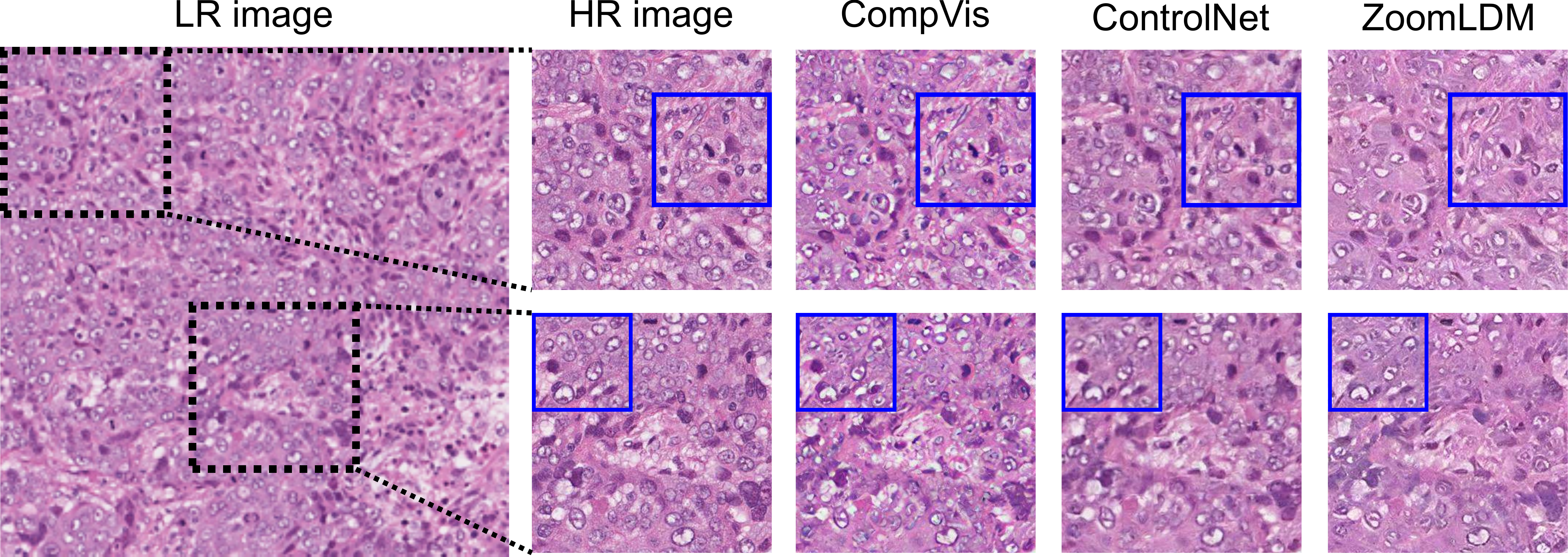}
    \centering
    \caption{We showcase $4 \times$ super-resolution results ($256 \times 256 \rightarrow 1024 \times 1024$). Samples generated by other methods \cite{rombach2022high, zhang2023adding} exhibit artifacts, inconsistencies, and blurriness that are not present in our outputs. Specifically, in blue boxes, we can observe that CompVis\cite{rombach2022high}  generates fine scale artifacts, while ControlNet\cite{zhang2023adding} produces generally blurry outputs. ZoomLDM produces a sharp output, generating details generally consistent with the ground truth image.}
    \label{fig:sr}
\end{figure*}

\subsection{Image quality}
For every histopathology magnification, we generate 10,000 $256 \times 256$ px patches using ZoomLDM and evaluate their quality using the Fréchet Inception Distance (FID) \cite{heusel2017gans}. For $20\times$, $10\times$ and $5\times$ magnifications, we compare against the state-of-the-art (SoTA) works of \cite{graikos2024learned,Yellapragada_2024_WACV}. For lower magnifications, we train standalone models specifically for patches from those magnifications, keeping the architecture consistent with ZoomLDM.

As indicated in Table~\ref{tab:fid}, ZoomLDM achieves superior performance across all magnifications compared to the SoTA models. We see larger improvements for magnifications below $2.5\times$, where the data scarcity severely impacts the model's ability to synthesize diverse, high-quality images. This highlights the advantage of our unified architecture and conditioning approach. By leveraging data and conditioning across all magnifications, we allow the low-density data regions to benefit from the insights that the model gains from the entire dataset, improving both model performance and efficiency.

\noindent \textbf{Novel image synthesis:} For FID comparisons above, images were generated by randomly sampling SSL embeddings for different magnifications from the dataset. However, this approach is not always practical as it requires access to the dataset of embeddings at all times. To address this, we use the Conditioning Diffusion Model to draw samples from the shared cross-magnification latent space and generate new images conditioned on these latents (CDM row in Table~\ref{tab:fid}). Despite the slight increase in FID -- an expected outcome since the CDM cannot perfectly capture the true learned conditioning latent space, we still observe that the generated samples outperform the baselines in the data-scarce settings. We believe that this further emphasizes the importance of our shared cross-magnification latent space, by showing that we can model its distribution and capture all scales effectively. In supplementary we show synthetic images at $0.15625\times$ and with their closest neighbors in the dataset to demonstrate the absence of memorization.

\begin{table}[ht]
\caption{CLIP and Crop FID values (lower is better) for our large image generation experiments. ZoomLDM outperforms previous works on $1024 \times 1024$ generation. While we lack in $4096 \times 4096$ FIDs, we provide qualitative examples in the supplementary that highlight the fundamental differences that emerge when scaling up the three methods. Inference time for a single image shows that our method is the only practical approach for 4k image generation.}
\addtolength{\tabcolsep}{-0.4em}
\begin{tabular}{|c||c|cc||c|cc|}
\hline
\multirow{2}{*}{Method} & \multicolumn{3}{c||}{$1024 \times 1024$}                                                                                                                                     & \multicolumn{3}{c|}{$4096 \times 4096$}                                                                                                                                      \\ \cline{2-7} 
                        & \begin{tabular}[c]{@{}c@{}}Time\\ / img\end{tabular} & \begin{tabular}[c]{@{}c@{}}CLIP\\  FID\end{tabular} & \begin{tabular}[c]{@{}c@{}}Crop \\ FID\end{tabular} & \begin{tabular}[c]{@{}c@{}}Time\\ / img \end{tabular} & \begin{tabular}[c]{@{}c@{}}CLIP\\  FID\end{tabular} & \begin{tabular}[c]{@{}c@{}}Crop \\ FID\end{tabular} \\ \hline
\citet{graikos2024learned}            & 60 s                                                 & 7.43                                                & 15.51                                               & 4 h                                                   & 2.75                                                & \textbf{11.30}                                      \\
$\infty$-Brush  \cite{le2024inftybrushcontrollablelargeimage}           & 30 s                                                 & 3.74                                                & 17.87                                               & 12 h                                                  & \textbf{2.63}                                       & 14.76                                               \\
ZoomLDM                 & 28 s                                                 & \textbf{1.23}                                       & \textbf{14.94}                                      & 8 m                                                  & 6.75                                                & 18.90                                               \\ \hline
\end{tabular}
\label{tab:large_image}
\end{table}

\begin{table*}[ht]
\centering
\small
\caption{Super-resolution results on TCGA-BRCA \cite{cancer2013cancer} and BACH \cite{ARESTA2019122} using ZoomLDM and other diffusion-based baselines. Using ZoomLDM with the proposed condition inference achieves the best performance.}
\addtolength{\tabcolsep}{-0.35em}
\begin{tabular}{l|c|ccccc|ccccc}
\hline
\multirow{2}{*}{Method}  & \multirow{2}{*}{Conditioning} & \multicolumn{5}{c|}{TCGA BRCA}                                                      & \multicolumn{5}{c}{BACH}                                                            \\ \cline{3-12} 
                         &                               & SSIM  $\uparrow$         & PSNR   $\uparrow$         & LPIPS$\downarrow$          & CONCH $\uparrow$         & UNI  $\uparrow$          & SSIM  $\uparrow$        & PSNR $\uparrow$          & LPIPS$\downarrow$         & CONCH  $\uparrow$        & UNI $\uparrow$           \\ \hline
Bicubic                 & -                             & \textbf{0.653} & \textbf{24.370} & 0.486          & 0.871          & 0.524          & \textbf{0.895} & \textbf{34.690} & \underline{0.180}          & 0.969          & \textbf{0.810 }         \\
CompVis \cite{rombach2022high}                  & LR image                      & 0.563          & 21.926          & \underline{0.247}          & 0.946          & 0.565          & 0.723          & 27.278          & 0.206          & 0.954          & 0.576          \\
ControlNet \cite{zhang2023adding}               & LR image                & 0.543          & 21.980          & 0.252          & 0.874          & 0.563          & 0.780              & 27.339               & 0.276              & 0.926              & 0.721              \\ \hline
\multirow{3}{*}{ZoomLDM} & Uncond                 & 0.591          & 23.217          & 0.260          & 0.936          & \underline{0.680}          & 0.739          & 29.822          & 0.235          & 0.965          & 0.741          \\
                         & GT emb                      & 0.599          & 23.273          & 0.250          & \underline{0.946}          & 0.672          & 0.732          & 29.236          & 0.245          & \underline{0.974}          & 0.753          \\
                         & Infer emb                & \underline{0.609}          & \underline{23.407}          & \textbf{0.229} & \textbf{0.957} & \textbf{0.719} & \underline{0.779}          & \underline{30.443}          & \textbf{0.173} & \textbf{0.974} & \underline{0.808} \\ \hline
\end{tabular}

\label{tab:sr}
\end{table*}
 
\subsection{Large image generation}
In Section~\ref{subsec:joint_sampling}, we presented an algorithm for jointly sampling images at multiple scales. We perform experiments on generating $20\times$ histopathology images jointly with other magnifications in two settings: Sampling $20\times$ with $5\times$, generating $1024 \times 1024$ images and sampling $20\times$ with $1.25\times$, giving $4096 \times 4096$ samples. We employ bicubic interpolation as the downsampling operator $\mA$, where for $5\times$ and $1.25\times$, we downsample by $4\times$ and $16\times$, respectively.

In Table~\ref{tab:large_image}, we showcase CLIP FID and Crop FID values, adopted from \cite{le2024inftybrushcontrollablelargeimage}, and compare our large-image generation method against existing state-of-the-art approaches. CLIP FID downsamples the full image and extracts features from a CLIP \cite{radford2021learning} model, whereas Crop FID extracts $256 \times 256$ crops from the large images and computes FID using the conventional Inception features \cite{Seitzer2020FID}.

On $1024 \times 1024$ generation we easily outperform existing approaches with similar or smaller sampling times. While, on $4096 \times 4096$ generation, we find that our method lags in two quality metrics but offers a reasonable inference time per image (8min vs $>4$hrs). However, regarding the $4096 \times 4096$ results, we find fundamental differences between our synthesized images (Figure~\ref{fig:large-image}) and those of \cite{graikos2024learned,le2024inftybrushcontrollablelargeimage} (see supplementary). We particularly find that the local patch-based model of \citet{graikos2024learned} completely fails to capture the global context in the generated images. While it generates great quality patches and stitches them together over the $4096 \times 4096$ canvas, the overall image does not resemble a realistic pathology image. On the other hand, $\infty$-Brush \cite{le2024inftybrushcontrollablelargeimage} captures the global image structures but produces blurry results. In contrast, ZoomLDM balances local details and global structure, producing images that not only exhibit high fidelity but also maintain overall realism across the entire $4096 \times 4096$ canvas. We are the first to generate 4k pathology images with both detail and global coherency under a tractable computational budget.

\subsection{Super-resolution}
Our joint multi-scale sampling allows us to sample multiple images from different magnifications simultaneously. However, a question arises of whether we could also use ZoomLDM in super-resolution, where the higher-scale image is given and the details need to be inferred. We provide a solution for super-resolution with ZoomLDM using a straightforward extension of our joint sampling algorithm.

The main challenge we need to overcome is the absence of conditioning. Given only an image at a magnification other than $20\times$, we cannot obtain SSL embeddings, which are extracted from a $20\times$-specific encoder. Nevertheless, we discover an interesting inversion property of our model, which allows us to infer the conditioning given an image and its magnification. Similar to textual inversion \cite{galimage}, and more recently prompt tuning \cite{chungprompt}, we can optimize the SSL input to the summarizer to obtain a set of embeddings that generate images that resemble the one provided. We discuss the inversion approach in the supplementary material in more detail, along with inversion examples.

Once we have obtained a set of plausible conditioning embeddings, we can run our joint multi-scale sampling algorithm, fixing the measurement $\vy$ to the real image we want to super-resolve. To test ZoomLDM's capabilities, we construct a simple testbed of $4\times$ super-resolution on in-distribution and out-of-distribution images from TCGA-BRCA and BACH \cite{ARESTA2019122} respectively. As baselines, we use bicubic interpolation, a naive super-resolution-specific LDM trained on OpenImages \cite{OpenImages} (CompVis), and a ControlNet \cite{zhang2023adding} trained on top of ZoomLDM.

In Table~\ref{tab:sr} and Figure~\ref{fig:sr}, we present the results of our experiments. We find that SSIM and PSNR are slightly misleading as they favor the blurry bicubic images, but also point out some significant inconsistencies in the LDM and the ControlNet outputs. For better comparisons, we also compute LPIPS \cite{zhang2018perceptual} and CONCH \cite{lu2024avisionlanguage} similarity, which downsamples the image to $224 \times 224$ as well as UNI similarity, which we consider on a per $256\times 256$ patch-level. In most perceptual metrics, we find ZoomLDM inference to be the best-performing while remaining faithful to the input image. Interestingly, we discover that using the embedding inversion that infers the conditions from the low-res given image performs better than providing the real embeddings.

\begin{table}[ht]
\centering
\caption{AUC for BRCA subtyping and HRD prediction. Features extracted from ZoomLDM outperform SoTA vision encoders.}
\begin{tabular}{c|c|cc}
\hline
Features        & Mag                                                    & Subtyping      & HRD            \\ \hline
Phikon \cite{filiot2023scaling}                  & $20\times$                                                              & 93.81          & 76.88          \\
UNI \cite{chen2023general}                      & $20\times$                                                              & 94.09          & 81.79          \\
CTransPath \cite{wang2021transpath}               & $5\times$                                                               & 93.11          & 85.37          \\ \hline
\multirow{3}{*}{ZoomLDM} & $20\times$                                                              & 94.49          & 85.25          \\
                         & $5\times$                                                               & 94.09          & 86.26          \\
                         & \begin{tabular}[c]{@{}c@{}}Multi-scale \\ ($20\times$ + $5\times$)\end{tabular} & \textbf{94.91} & \textbf{88.03} \\ \hline
\end{tabular}
\label{tab:mil}
\end{table}

\subsection{Multiple Instance Learning}

Multiple instance learning (MIL) tasks benefit from multi-scale information, as different magnifications reveal distinct and complementary features. Prior work \cite{li2021dual, chen2022scaling} that demonstrated this behavior required training separate encoders for each scale. We hypothesize that ZoomLDM offers an efficient solution by enabling seamless multi-scale feature extraction. 
 
To validate this hypothesis, we utilize ZoomLDM as a feature extractor and apply a MIL approach for slide-level classification tasks of Breast cancer subtyping and Homologous Recombination Deficiency (HRD) prediction - both of which are binary classification tasks. For each patch in the WSI, we extract features from ZoomLDM's U-Net output block 3 at a fixed timestep $t=100$, conditioned on UNI embeddings. We employ a 10-fold cross-validation strategy for subtyping, consistent with the data splits from HIPT \cite{chen2022scaling}, and a 5-fold cross-validation for HRD prediction, reporting performance on a held-out test split as per SI-MIL \cite{kapse2024si}. We compare ZoomLDM’s features to those from SoTA encoders—Phikon \cite{filiot2023scaling}, CTransPath \cite{wang2021transpath}, and UNI \cite{chen2023general}, using the ABMIL method \cite{ilse2018attention, kaczmarzyk2024explainable}.

As expected, the results in Table \ref{tab:mil} show that ZoomLDM's multi-scale features (fusing $20\times$ and $5 \times$ outperform SoTA encoders in both tasks. This improvement highlights the effectiveness of ZoomLDM's cross-magnification latent space in capturing multi-scale dependencies. Surprisingly, even in a single magnification setting, ZoomLDM outperforms all SoTA encoders.
This result suggests that by learning to generate across scales, ZoomLDM learns to produce features that can be aware of the cross-magnification long-range dependencies, and therefore exceed the capabilities of those produced by SSL encoders for downstream MIL tasks.

\section{Conclusion}
We presented ZoomLDM, the first conditional diffusion model capable of generating images across multiple scales with state-of-the-art synthetic image quality. By introducing a cross-magnification latent space, implemented with a trainable summarizer module, ZoomLDM effectively captures dependencies across magnifications. Our novel joint multi-scale sampling approach allows for efficient generation of large, high-quality and structurally coherent histopathology images up-to $4096 \times 4096$ pixels while preserving both global structure and fine details. 

In addition to synthesis, ZoomLDM demonstrates its utility as a powerful feature extractor in multiple instance learning experiments. The multi-scale representations learned by our model outperform SoTA SSL encoders in slide-level classification tasks, enabling more accurate subtyping, prognosis prediction, and biomarker identification. Furthermore, our Condition Diffusion Model demonstrates the potential to integrate diverse input sources such as text or RNA sequences, paving the way for realistic synthetic datasets for training and evaluating pathologists as well as controlled datasets for quality assurance. ZoomLDM is a step toward achieving foundation generative models in histopathology, with the potential to shed light on tumor heterogeneity, refine cancer gradings, and enrich our understanding of cancer's various manifestations.

\paragraph{Acknowledgements} This research was partially supported by NSF grants IIS-2123920, IIS-2212046, NIH grants 1R01CA297843-01, 3R21CA258493-02S1 and NCI awards 1R21CA25849301A1, UH3CA225021.

{
    \small
    \bibliographystyle{ieeenat_fullname}
    \bibliography{main}
}

\maketitlesupplementarysingle

\noindent We organize the supplementary as follows:
\begin{itemize}
\item[] 
\begin{itemize}
    \item[\ref{sec:supp_satellite}] ZoomLDM on satellite images
    \item[\ref{sec:supp_ablation}] Ablation on SSL encoder and Summarizer
    \item[\ref{sec:supp_experiments}] Experiment details:
    \begin{itemize}
        \item[\ref{subsec:supp_summarizer_cdm}] Summerizer-CDM training details
        \item[\ref{subsec:supp_joint_sampling}] Joint sampling
        \item[\ref{subsec:supp_inversion}] Image inversion
    \end{itemize}
    \item[\ref{sec:supp_additional}] Additional Details
    \begin{itemize}
        \item[\ref{subsec:supp_superres}] More super-resolution baselines
        \item[\ref{subsec:supp_memorization}] Data efficiency and memorization
        \item[\ref{subsec:supp_patches}] Patches from all scales
        \item[\ref{subsec:supp_large_images}] Generated large images
        \item[\ref{subsec:supp_prev_works}] Comparison to previous works
    \end{itemize}
\end{itemize}
\end{itemize}

\section{ZoomLDM on satellite images}
\label{sec:supp_satellite}
In the main text, we focused on the digital histopathology domain and how our multi-scale diffusion model can prove useful in generation and downstream tasks. However, gigapixel images also concern the remote sensing domain, where satellite images regularly are in the range of $10000 \times 10000$ pixels. To show the wide applicability of our multi-scale approach, we trained ZoomLDM on satellite images from the NAIP dataset \cite{naip}, specifically using NAIP images from the Chesapeake subset of \cite{robinson2019large}. NAIP images are at 1m resolution -- the distance between pixel centers is 1m. We follow the same dataset preparation approach and extract $256 \times 256$ patches at four different scales with pixels corresponding to 1m, 2m, 4m, and 8m resolutions. For the SSL encoder, we resort to a pre-trained DINOv2 model \cite{oquab2023dinov2}, which has been known to perform well across many modalities, including satellite.

In Table~\ref{tab:naip_fid}, we provide the per-resolution FID numbers our model achieves. Similarly to histopathology, we observe that training a cross-scale model benefits the scales where there is not enough data to train a single-scale model on (8m resolution in this case). We also showcase patches generated by ZoomLDM at all four resolutions in Figure \ref{fig:naip_all_mag}. We present examples from the satellite ZoomLDM variant in \ref{subsec:supp_joint_sampling} and \ref{subsec:supp_patches}.

In Table~\ref{tab:naip_large_fid}, we provide the FID numbers for large satellite image generation ($1024\times1024$). Our satellite ZoomLDM model achieves significantly better results on crop FID while achieving similar CLIP FID; this showcases our ability to synthesize high-quality images that simultaneously maintain global consistency.

\begin{table*}[ht]
\centering
\begin{minipage}[t]{0.6\linewidth}
    \centering
    \begin{tabular}{|c|cccc|}
    \hline
    Resolution          & 1m             & 2m            & 4m            & 8m            \\ \hline
    \# Training patches & 365 k          & 94 k          & 25 k          & 8.7 k         \\ \hline
    ZoomLDM             & \textbf{10.93} & \textbf{7.77} & \textbf{7.34} & \textbf{8.46} \\
    SoTA model          & 11.5 \cite{graikos2024learned}           & 23.61         & 37.52         & 65.45         \\ \hline
    \end{tabular}
    \caption{NAIP FID values obtained by ZoomLDM versus training a state-of-the-art diffusion model on a single resolution. Having a shared model across multiple scales improves the generation quality for the data-scarce scales. For resolutions \textgreater 1m we retrain the model of \cite{graikos2024learned} on the samples from that resolution only.}
    \label{tab:naip_fid}
\end{minipage}
\hfill
\begin{minipage}[t]{0.35\linewidth}
    \centering
    \begin{tabular}{|c|cc|}
    \hline
    \multirow{2}{*}{Method}  & CLIP & Crop \\
     & FID & FID  \\
    \hline
    Graikos et al. \cite{graikos2024learned} & 6.86 & 43.76 \\
    $\infty$-Brush \cite{le2024inftybrushcontrollablelargeimage} & \textbf{6.32} & 48.65 \\
    ZoomLDM  & 7.90 & \textbf{13.25}  \\
    \hline
    \end{tabular}
    \caption{CLIP and Crop FID values (lower is better) for large ($1024\times1024$) satellite images. ZoomLDM outperforms previous works while also maintaining a reasonable inference time.}
    \label{tab:naip_large_fid}
\end{minipage}
\end{table*}

\section{Ablation on SSL encoder and Summarizer}
\label{sec:supp_ablation}
We retrain ZoomLDM with (i) a weaker SSL encoder (HIPT \cite{chen2022scaling}) and (ii) both a weaker SSL encoder and a simpler summarizer network (CNN vs ViT). Table~\ref{tab:ssl_summ_ablation} shows that replacing UNI with HIPT degrades performance and further replacing the ViT summarizer network with a simple 4-layer CNN leads to a greater decline. 

When comparing the downstream performance of the denoiser features on a multiple-instance learning task (MIL) we also see a decrease in performance when using a 'weaker' conditioning encoder. We believe that training a diffusion model conditioned on SSL representations complements the discriminative SSL pre-training with the newly learned generative features. In all our experiments, improved image quality leads to better downstream task performance. Additionally, the SSL encoders used in MIL are usually trained on a single magnification, making our approach a potential way to fuse features across different scales effectively.

\begin{table*}[t]
\centering
\small
\begin{tabular}{|c|c|cccccccc|cc|}
\hline
\multirow{2}{*}{SSL} & \multirow{2}{*}{Summarizer} & \multicolumn{8}{c|}{FID across magnifications \textdownarrow} & \multicolumn{2}{c|}{MIL (AUC) \textuparrow} \\
                     &                             & $20\times$ & $10\times$ & $5\times$ & $2.5\times$ & $1.25\times$ & $0.625\times$ & $0.3125\times$ & $0.15625\times$ & Subtyping & HRD \\ \hline
HIPT \cite{chen2022scaling} & CNN & 18.88 & 16.75 & 19.31 & 16.01 & 14.45 & 14.21 & 15.44 & 18.47 & 86.20 & 72.44 \\
HIPT \cite{chen2022scaling} & ViT & 13.49 & 14.42 & 15.84 & 13.32 & 14.32 & 12.31 & 16.25 & 19.90 & 87.26 & 75.92 \\
UNI \cite{chen2024uni}      & ViT & \textbf{6.77} & \textbf{7.60} & \textbf{7.98} & \textbf{10.73} & \textbf{8.74} & \textbf{7.99} & \textbf{8.34} & \textbf{13.42} & \textbf{94.49} & \textbf{85.25} \\ 
\hline
\end{tabular}
\caption{Ablation on SSL encoder and summarizer network architecture. Using a weaker SSL encoder or summarizer leads to worse performance in both generation and downstream discriminative tasks.}
\label{tab:ssl_summ_ablation}
\end{table*}

\section{Experiment details}
\label{sec:supp_experiments}

\subsection{Summarizer-CDM training details}
\label{subsec:supp_summarizer_cdm}

\noindent\textbf{Summarizer:} We train the Summarizer jointly with the LDM. The Summarizer processes the SSL embeddings extracted alongside the image patches and projects them to a latent space that is shared across all scales (cross-magnification latent space). By training jointly with the LDM the Summarizer learns to compress the SSL embeddings into a representation useful for making images.

We pre-process the SSL embedding matrices via element-wise normalization. The Summarizer receives 64 SSL embeddings (or fewer SSL embeddings with appropriate padding to 64 tokens) concatenated with a learned magnification embedding as input. The network consists of a 12-layer Transformer encoder with a hidden dimension of 512, followed by a LayerNorm operation to normalize the output. The $65 \times 512$ dimensional output is then fed to the U-Net denoiser via cross-attention.

\begin{figure}[ht]
    \includegraphics[width=\textwidth]{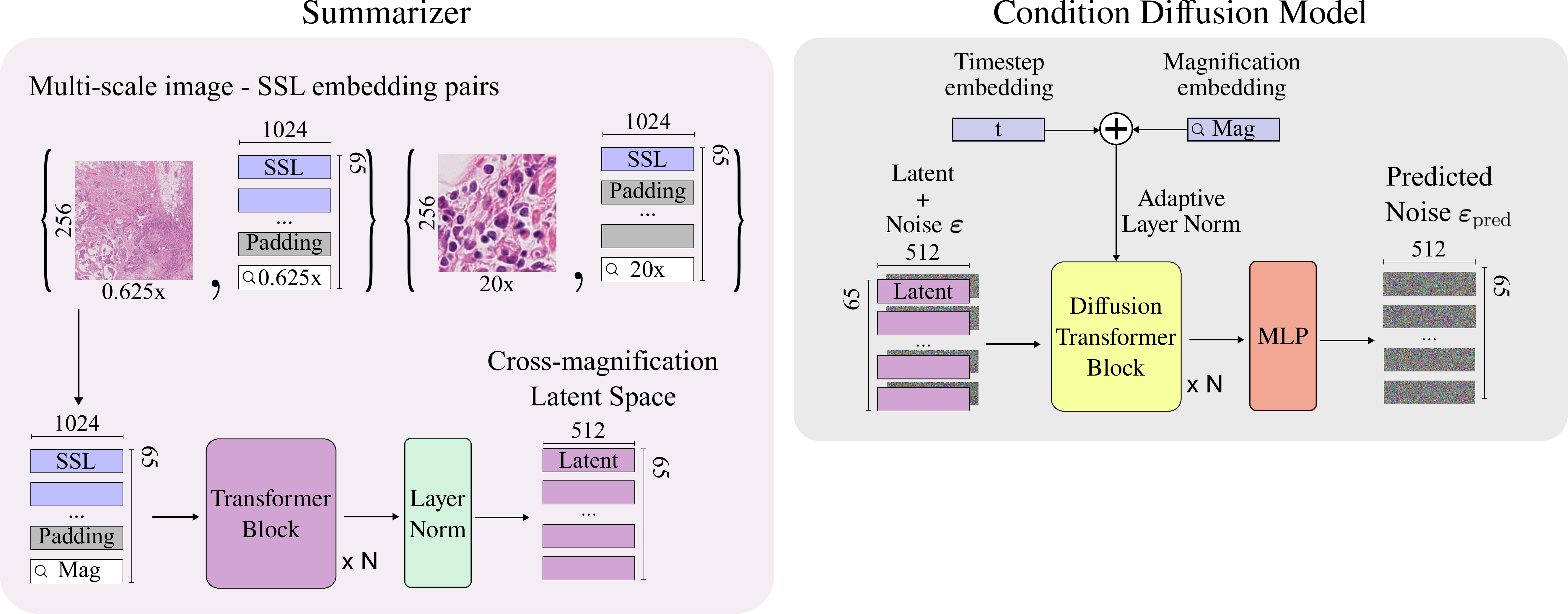}
    \centering
    \caption{Overview of the Summarizer and Condition Diffusion Model.}
    \label{fig:summarizer_cdm}
\end{figure}

\noindent\textbf{CDM:} To avoid reliance on real images to extract the SSL embeddings required for sampling, we train a Conditioning Diffusion Model (CDM). The CDM is trained to draw samples from the learned cross-magnification latent space. After training the LDM and Summarizer jointly, we train the CDM with the denoising objective to sample from the $65 \times 512$ output. See Figure \ref{fig:summarizer_cdm} for an overview of the Summarizer and CDM.

\begin{figure}[ht]
    \includegraphics[width=\textwidth]{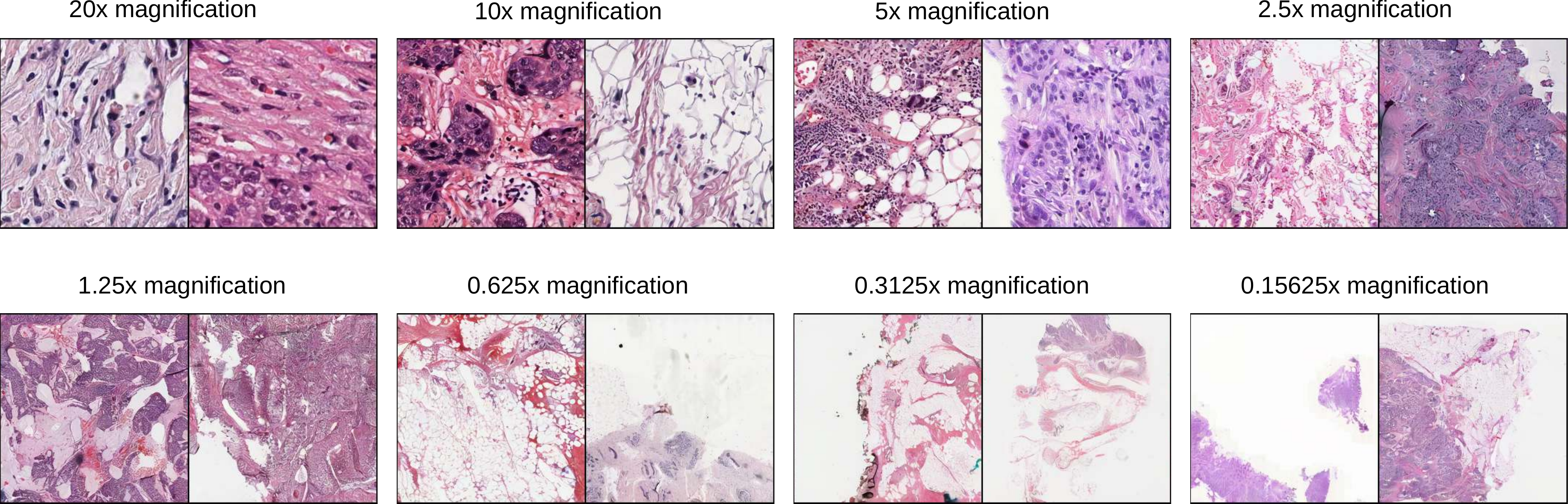}
    \centering
    \caption{Images synthesized by ZoomLDM using conditions sampled from our Conditioning Diffusion model (CDM).}
    \label{fig:cdm_samples}
\end{figure}

We implement the CDM as a Diffusion Transformer \cite{peebles2023scalable}. We use the DiT-Base architecture, consisting of 12 layers and a hidden size of 768. We use an MLP to project the output back to the exact channel dimensions as the input. We use a constant learning rate of $10^{-4}$, following the implementation of \cite{peebles2023scalable}.  We present samples generated by the CDM in Figure \ref{fig:cdm_samples}.

\subsection{Joint Sampling}
\label{subsec:supp_joint_sampling}

In this section, we present an overview of the joint sampling algorithm. By jointly generating an image that depicts the global context and images that produce local details we are able to synthesize large images at the highest resolution that maintain global coherency. We achieve that by simultaneously generating patches $i$ with high-resolution details $\vx^i = \dec(\vz^i)$ and a lower-resolution context $\vx^L = \dec(\vz^L)$ that globally guides the structure of the patches.

Our joint sampling method is based on a recent fast sampling algorithm for diffusion models under linear constraints, presented in \cite{graikos2024fast}. The full algorithm is shown in Algorithm~\ref{alg:sampling_backprop}. We make two key changes to the inference algorithm to perform joint multi-scale sampling: (i) We replace the constraint $\vy$ with the current estimate of the lower-scale image $\dec(\hat{\vz}_0^L)$ and (ii) we replace the expensive backpropagation step required in computing the error $\ve$ with a less memory-intensive approximation using forward passes through the encoder and the decoder.

\textbf{Utilizing intermediate steps} Instead of having access to a measurement $\vy$ we only have access to the current estimate of the context image. That image is in practice a subsampled version of the spatially arranged patches $\vx^i$. To relate the two, we rearrange $\vx_i$ and apply a linear subsampling operator $\mA$, such as bicubic interpolation. This operator is used to compute the difference between the current synthesized patches and the current context and will be used to update the content of the patch images.

\textbf{Avoiding backpropagation} For latent diffusion models, the original algorithm relies on computing the difference between the context and the patches which it then backpropagates through the decoder to get the direction towards which this error is minimized. However, when we synthesize 4k images, we end up with 256 high-resolution patches, and backpropagating becomes prohibitively memory-intensive. To that end, we propose a modification to the sampling algorithm that replaces the backpropagation step with forward passes through the encoder and decoder.

To produce the high-resolution images, we want to sample $z_t$ under the guidance of the lower-scale image, minimizing a constraint $C(\vz_t) = || \mA \dec(\hat{\vz}_0(\vz_t)) - \dec(\hat{\vz}_0^L) ||_2^2$. Algorithm~\ref{alg:sampling_backprop} requires us to compute the direction $\ve$ of $\hat{\vz}_0$ towards which the constraint $C$ is minimized and uses it to update the current diffusion latent as
\begin{gather}
    \vg = \frac{\hat \vz_0(\vz_t + \delta \ve) - \hat \vz_0 (\vz_t)}{\delta} \\
    z_t' = z_t + \lambda g.
\end{gather}
However, to calculate $\vg$ we need $\ve = \frac{\partial C}{\partial \hat \vz_0}$ which we can calculate by backpropagating through the decoder model. Since this is computationally burdensome, we apply the chain rule to get
\begin{equation}
\label{eq:J1}
    \ve = \frac{\partial C}{\partial \hat \vz_0 } = \left(\frac{\partial \dec(\hat \vz_0 )}{\partial \hat \vz_0 }\right)^T\frac{\partial C}{\partial \dec(\hat \vz_0 )} = \left(\frac{\partial \dec(\hat \vz_0 )}{\partial \hat \vz_0 }\right)^T\ve_{\textit{img}},
    \quad \ve_{\textit{img}} = \mA^T(\mA \dec(\hat \vz_0(\vz_t)) - \dec(\hat \vz_0^L) )
\end{equation}
The LDM VAEs that we use (VQ-VAE or KL-VAE) are trained in a way that forces the Jacobian of the Decoder to be approximately orthogonal, through vector quantization or minimizing the KL divergence between the predicted posterior and an isotropic Gaussian. For orthogonal Jacobians Eq.~\ref{eq:J1} can be simplified into:
\begin{equation}
\label{eq:J2}
    \ve = \left(\frac{\partial \dec(\hat \vz_0)}{\partial \hat \vz_0}\right)^Te_{\textit{img}} \approx \frac{\partial \hat \vz_0}{\partial \dec(\hat \vz_0)}e_{\textit{img}} 
\end{equation}
and assuming that the VAE has learned to reconstruct images perfectly, it can be written as:
\begin{equation}
\label{eq:J3}
    \ve \approx \frac{\partial \hat \vz_0}{\partial \dec(\hat \vz_0)}e_{\textit{img}} \approx \frac{\partial \enc(\dec(\hat \vz_0))}{\partial \dec(\hat \vz_0)}e_{\textit{img}}.
\end{equation}
We can now approximate $\ve$ using finite differences:
\begin{equation}
\label{eq:J4}
    \ve \approx \frac{\partial \enc(\dec(\hat \vz_0))}{\partial \dec(\hat \vz_0)}e_{\textit{img}}  \approx \frac{\enc(\dec(\hat \vz_0) + \zeta e_{\textit{img}}) - \enc(\dec(\hat \vz_0))}{\zeta}
\end{equation}
which completely erases the need to perform memory-heavy backpropagation through the decoder model. 

\begin{figure}[hb]
    \includegraphics[width=\textwidth]{figures/multi_scale_examples_brca.pdf}
    \centering
    \caption{Joint sampling process across two different magnifications for the TCGA-BRCA ZoomLDM model. We jointly generate a $256 \times 256$ image at $1.25\times$ and a $4096 \times 4096$ image at $20\times$. The $1.25\times$ generation guides the structure of the $20\times$ image by providing the necessary global context that each $20\times$ patch is unaware of. The generated large $20\times$ image has a realistic global arrangement of cells and tissue. Best viewed zoomed-in.}
    \label{fig:multi_scale_brca}
    
    \includegraphics[width=\textwidth]{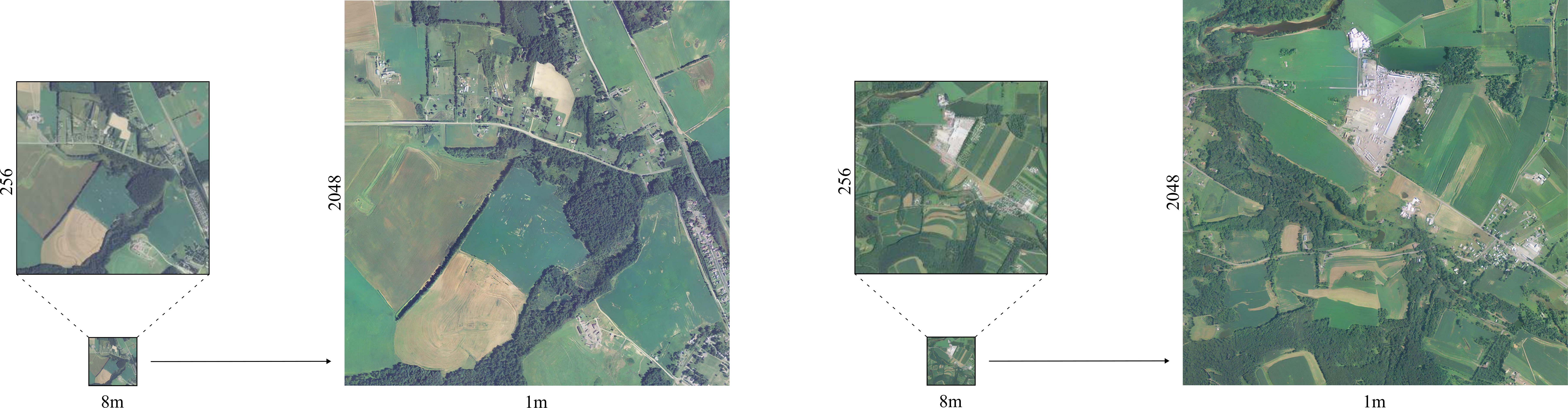}
    \centering
    \caption{Joint sampling process across two different resolutions for the Satellite ZoomLDM model. We jointly generate a $256 \times 256$ image at $8m$ resolution and a $2048 \times 2048$ image at $1m$. The $8m$ generation guides the structure of the $1m$ image by providing global coherence, which, otherwise, each $1m$ would be unaware of. The generated large $1m$ image has realistic global structures, with roads and forests neatly arranged across the $2048 \times 2048$ canvas. Best viewed zoomed-in.}
    \label{fig:multi_scale_naip}
\end{figure}

A step-by-step description of our joint sampling method can be found in Algorithm~\ref{alg:sampling_vae}. We use 50 DDIM steps for our experiments, bicubic upsampling/downsampling for $\mA$, $\delta = \zeta = 0.005$, $K=1$, $\lambda=0.5$. Upon observing noticeable discontinuities along the borders of the high-resolution patches, we apply a simple post-processing step by adding noise and denoising the patches between, similar to \cite{graikos2024learned}. We provide some results of the joint sampling, visualized in Figures~\ref{fig:multi_scale_brca},\ref{fig:multi_scale_naip} for the histopathology and satellite domains.

\begin{minipage}[t]{0.4\textwidth}
\centering
\begin{algorithm}[H]
\caption{The algorithm for linear inverse problem solving proposed in \cite{graikos2024fast}.}
\label{alg:sampling_backprop}
\begin{algorithmic}
    \State \textbf{Input:} Diffusion model $\hat{\vz}_0(\vz_t),\ \enc,\ \dec$, schedule $T_{0,\dots,M}$, subsampling operator $\mA$, measurement $\vy$, step size $\delta$, \# iterations $K$, learning rate $\lambda$
    \State $\vz_T \sim N(\textbf{0}, \mI)$
        \For{$t \in \{T_0, T_1,\dots,T_M\}$}
            \For{$i \in \{1, 2,\dots,K\}$}
                \State $\ve = \nabla_{\vz_0} || \mA \dec(\hat{\vz}_0(\vz_t)) - \vy ||_2^2$
                \State $\vg = \left[ \hat{\vz}_0(\vz_t+\delta \ve)-\hat{\vz}_0(\vz_t) \right]/\delta$
                \State $\vz_t = \vz_t + \lambda \vg$
            \EndFor
            \State $\vz_{t} = \text{DDIM}(\vz_t, \hat{\vx}_0, s)$
        \EndFor
    \State \textbf{Return:} $\vx_0$
\end{algorithmic}
\end{algorithm}
\end{minipage} %
\hfill
\begin{minipage}[t]{0.55\textwidth}
\centering
\begin{algorithm}[H]
\caption{The proposed modification to Algorithm~\ref{alg:sampling_backprop}.}
\label{alg:sampling_vae}
\begin{algorithmic}
    \State \textbf{Input:} Diffusion model $\hat{\vz}_0(\vz_t),\ \enc,\ \dec$, schedule $T_{0,\dots,M}$, subsampling operator $\mA$, detail scale $s$, context scale $s_L$, step sizes $\delta,\zeta$, \# iterations $K$, learning rate $\lambda$
    \State $\vz_T \sim N(\textbf{0}, \mI)$
    \State $\vz_T^L \sim N(\textbf{0}, \mI)$
        \For{$t \in \{T_0, T_1,\dots,T_M\}$}
            \For{$i \in \{1, 2,\dots,K\}$}
                \State $\ve_{\textit{img}} = \mA^T(\mA \dec(\hat{\vz}_0(\vz_t)) - \dec(\hat{\vz}_0^L) )$
                \State $\ve = \left[ \enc(\dec(\hat{\vz}_0) + \zeta \ve_{img}) - \enc(\dec(\hat{\vz}_0)) \right] / \zeta $
                \State $\vg = \left[ \hat{\vz}_0(\vz_t+\delta \ve)-\hat{\vz}_0(\vz_t) \right]/\delta$
                \State $\vz_t = \vz_t + \lambda \vg$
            \EndFor
            \State $\vz_{t} = \text{DDIM}(\vz_t, \hat{\vx}_0, s)$
            \State $\vz_{t}^L = \text{DDIM}(\vz_t^L, \hat{\vx}_0, s_L)$
        \EndFor
    \State \textbf{Return:} $\vx_0$
\end{algorithmic}
\end{algorithm}
\end{minipage}

\subsection{Image Inversion}
\label{subsec:supp_inversion}

In this section, we present our image inversion algorithm, which is crucial for performing the super-resolution task described in the main text. The conditioning we provide to the model is a set of SSL embeddings extracted at the highest resolution available. For instance, in histopathology, the SSL conditions are extracted at $20\times$. Thus, when we are given a single image at any magnification that we want to super-resolve we do not have access to this conditioning and are limited to using the model in an unconditional manner. The unconditional model is available since we randomly drop the conditioning during training, to implement classifier-free guidance \cite{ho2022classifier} during sampling. However, recent works have argued that when using the diffusion model to sample with linear constraints, like super-resolution, conditioning helps in achieving better-fidelity results \cite{chungprompt}.

\begin{figure}[ht]
    \includegraphics[width=\textwidth]{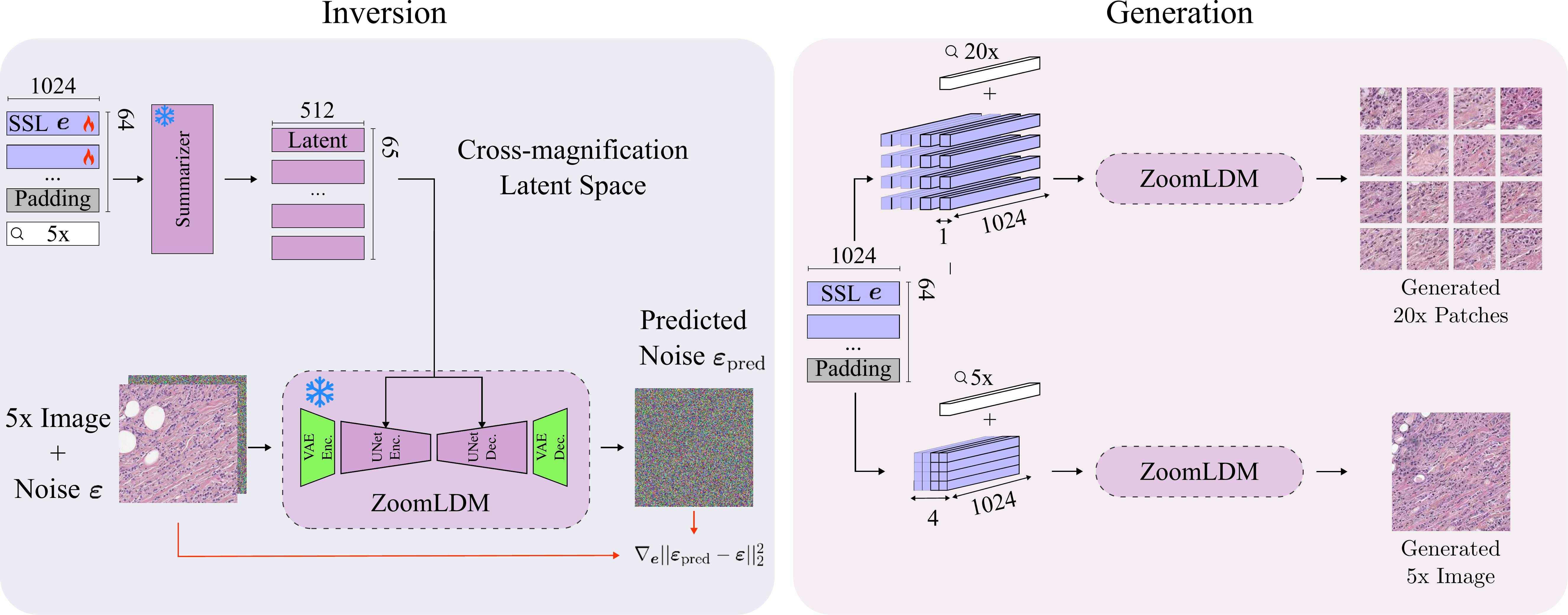}
    \centering
    \caption{Figure illustrating our pipeline for the image inversion used in the super-resolution task. For a given image we first use the denoising loss to optimize the input, conditioning embeddings. We can then generate variations of the given image and high-resolution patches from it. We use those per-patch embeddings to perform super-resolution, obtaining better results than unconditional super-resolution.}
    \label{fig:inversion}
\end{figure}

Inspired by those findings, we propose a simple algorithm to first \emph{invert} the model and get conditioning for a single image, before super-resolving it. The algorithm is an adaptation of the textual inversion technique of \citet{galimage}, which has seen wide success in text-to-image diffusion models. An overview of the approach is provided in Figure~\ref{fig:inversion}.

Given an image $\mI$ at scale $s$, we have access to a pre-trained latent denoiser model $\bm{\epsilon}_{\theta}(\vz_t, t, f(\ve, s))$ where $\vz = \enc(\mI)$, $g$ is the summarizer model and $\ve$ are the SSL embeddings that describe the image. We want to draw a sample $\ve$, that when provided as conditioning to the diffusion model will generate images similar to $\mI$. From the latent variable perspective of diffusion models, described by \citet{ho2020denoising}, we obtain the following lower bound for the log probability of $\vz$ given a condition $\ve$
\begin{equation}
    \log p(\vz \mid \ve) \geq -\sum_{t=1}^T w_t(\alpha) \mathbb{E}_{\bm{\epsilon}\sim \mathcal{N}(\bm{0},\mI)} \left[ || \bm{\epsilon_\theta}(\vz_t, t, g(\ve,s)) - \bm{\epsilon} ||_2^2 \right],
    \quad \vz_t = \sqrt{\alpha_t}\vz + \sqrt{1-\alpha}\bm{\epsilon}.
    \label{eq:elbo_ddpm}
\end{equation}

We then employ variational inference to fit an approximate posterior $q(\ve)$ to $p(\ve \mid \vz)$ from which we want to sample conditions given an input image. We start by defining a lower bound for $\log p(\vz)$
\begin{align}
    \log p(\vz) &= \log \int_{\ve} p(\vz, \ve) d\ve = \log \int_{\ve} q(\ve) \frac{p(\vz, \ve)}{q(\ve)} d\ve  \nonumber \\
    &= \log \mathbb{E}_{q(\ve)} \left[ \frac{p(\vz, \ve)}{q(\ve)} \right] \geq \mathbb{E}_{q(\ve)} \left[ \log \frac{p(\vz, \ve)}{q(\ve)} \right] \nonumber \\
    &= \mathbb{E}_{q(\ve)} \left[ \log \frac{p(\vz \mid \ve)p(\ve)}{q(\ve)} \right] = L.
    \label{eq:elbo_condition}
\end{align}
By maximizing the bound $L$ w.r.t. the parameters of $q$ we minimize the KL-Divergence between the approximate posterior $q(\ve)$ and the real $p(\ve \mid \vz)$. We choose a simple Dirac delta $q(\ve) = \delta (\ve - \vu)$ as our approximation, which allows us to use the bound from Eq.~\ref{eq:elbo_ddpm} to simplify the objective
\begin{align}
    L &= \mathbb{E}_{q(\ve)} \left[ \log p(\vz \mid \ve) + \log p(\ve) - \log q(\ve) \right] = \log p(\vz \mid \ve = \vu) + \log p(\ve = \vu) \nonumber \\
    &= -\sum_{t=1}^T w_t(\alpha) \mathbb{E}_{\bm{\epsilon}\sim \mathcal{N}(\bm{0},\mI)} \left[ || \bm{\epsilon_\theta}(\vz_t, t, g(\vu,s)) - \bm{\epsilon} ||_2^2 \right] + \log p(\ve = \vu).
    \label{eq:condition_opt}
\end{align}
Therefore, to draw a sample from the posterior $p(\ve \mid \vz)$ we optimize Eq.~\ref{eq:condition_opt} w.r.t. $\vu$. The result is a single point $\vu$ that seeks a local mode of $p(\ve \mid \vz)$. 

For the prior term $\log p(\ve)$, we use a simple heuristic, implementing a penalty that maximizes the similarity between the different vectors in the SSL embeddings $\ve$. This heuristic encourages the model to find embeddings that generate similar patches when used independently. For the denoising terms, we must add random Gaussian noise to the image latent $\vz$ and denoise at multiple timesteps $t$. Instead of evaluating multiple timesteps simultaneously, we utilize an annealing schedule that starts from $t=950$ and linearly decreases to $t=50$ over the $n=200$ optimization steps we perform. Overall, the proposed algorithm is similar to textual inversion \cite{galimage}, which utilizes the denoising loss to optimize text tokens $\vt$.

\begin{figure}[ht]
    \includegraphics[width=\textwidth]{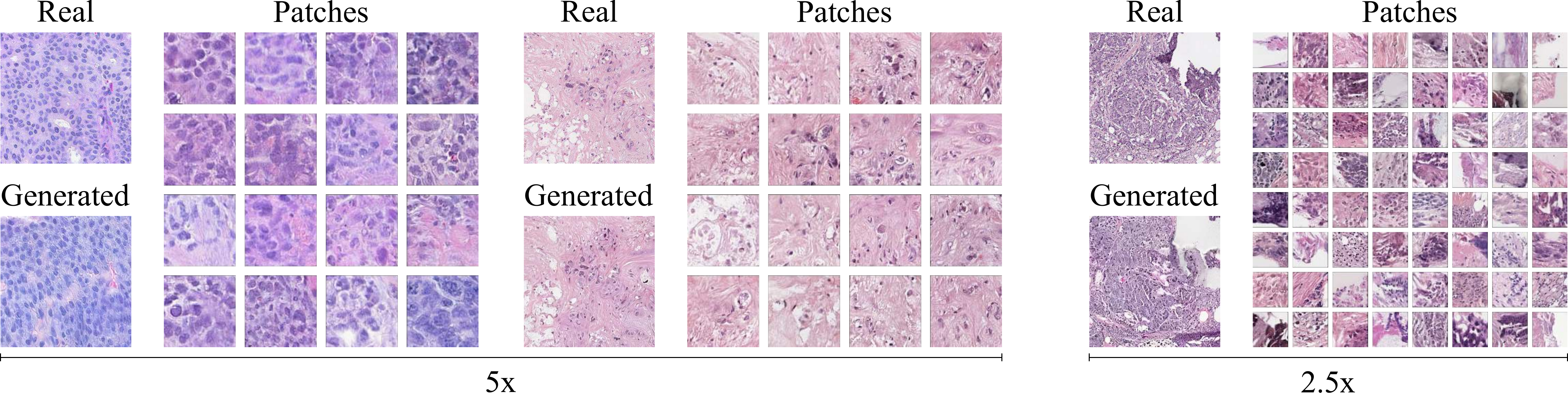}
    \centering
    \caption{Examples of the image inversion algorithm. Given a real image at any magnification, we infer the SSL embeddings that generated it. We then generate a new, similar-looking image at the same magnification using those embeddings as conditioning. Using the inferred embeddings to generate single patches from the given image yields convincing results at magnifications $>5\times$.}
    \label{fig:inversion_examples}
\end{figure}

In Figure~\ref{fig:inversion_examples}, we provide qualitative results for our inversion approach. We present two cases, inferring the condition for $5\times$ and $2.5\times$ images. We observe that for $5\times$, which is also the scale used in our super-resolution experiments, our approach can provide conditions that faithfully reconstruct both the $5\times$ image and also give us plausible $20\times$ patches. As we increase the number of conditions to infer, the $2.5\times$ result remains convincing at the lower scale but struggles to provide reasonable $20\times$ patches. Future work focusing on this inversion approach could provide useful insights into the SSL embeddings used as conditioning, helping understand what they encode and the topology of the latent space created by the SSL encoder.

\section{Additional results}
\label{sec:supp_additional}

\subsection{More super-resolution baselines}
\label{subsec:supp_superres}

In Tables~\ref{tab:supp_sr_tcga} and \ref{tab:supp_sr_bach} we provide additional baselines for the super-resolution task. We use ResShift \cite{yue2024resshift,yue2024efficient} and StableSR \cite{wang2024exploiting} to super-resolve pathology images and compare them to the zero-shot performance of ZoomLDM. Using ZoomLDM in a training-free manner (with condition inference \ref{subsec:supp_inversion}) remains the best approach for histopathology image super-resolution.

\begin{table}[ht]
\centering
\caption{Super-resolution results on TCGA-BRCA}
\addtolength{\tabcolsep}{-0.35em}
\begin{tabular}{l|ccccc}
\hline
Method & SSIM  $\uparrow$ & PSNR  $\uparrow$ & LPIPS$\downarrow$ & CONCH $\uparrow$ & UNI  $\uparrow$ \\ \hline
ResShift v2 (15 steps) \cite{yue2024resshift} & 0.415     & 19.716       & 0.431        & 0.847      & 0.299       \\ 
ResShift v3 (4 steps) \cite{yue2024efficient} & 0.525        & 21.528        & 0.314        & 0.866        & 0.311       \\ \hline
StableSR no tiling \cite{wang2024exploiting} & 0.515       & 21.644       & 0.315       & 0.862       & 0.390       \\ 
StableSR w/ tiling \cite{wang2024exploiting} & 0.514        & 21.618        & 0.316         & 0.863        & 0.388        \\ \hline
ZoomLDM (Uncond)     & 0.591          & 23.217          & 0.260          & 0.936          & \underline{0.680}          \\
ZoomLDM (GT Emb)     & 0.599          & 23.273          & 0.250          & \underline{0.946}          & 0.672          \\
ZoomLDM (Infer Emb)  & \underline{0.609} & \underline{23.407} & \textbf{0.229} & \textbf{0.957} & \textbf{0.719} \\ \hline
\end{tabular}
\label{tab:supp_sr_tcga}
\end{table}

\begin{table}[ht]
\centering
\caption{Super-resolution results on BACH}
\addtolength{\tabcolsep}{-0.35em}
\begin{tabular}{l|ccccc}
\hline
Method & SSIM  $\uparrow$ & PSNR $\uparrow$ & LPIPS$\downarrow$ & CONCH  $\uparrow$ & UNI $\uparrow$ \\ \hline
ResShift v2 (15 steps) \cite{yue2024resshift} & 0.584        & 23.256            & 0.421         & 0.898           & 0.621          \\ 
ResShift v3 (4 steps) \cite{yue2024efficient} & 0.751          & 26.283          & 0.257          & 0.898           & 0.623           \\ \hline
StableSR no tiling \cite{wang2024exploiting} & 0.729        & 26.203            & 0.291           & 0.846        & 0.547          \\ 
StableSR w/ tiling \cite{wang2024exploiting} & 0.729          & 26.200             & 0.293         & 0.845        & 0.538         \\ \hline
ZoomLDM (Uncond)        & 0.739          & 29.822          & 0.235          & 0.965          & 0.741          \\
ZoomLDM (GT Emb)        & 0.732          & 29.236          & 0.245          & \underline{0.974}          & 0.753          \\
ZoomLDM (Infer Emb)     & \underline{0.779} & \underline{30.443} & \textbf{0.173} & \textbf{0.974} & \underline{0.808} \\ \hline
\end{tabular}
\label{tab:supp_sr_bach}
\end{table}

\subsection{Data efficiency and memorization}
\label{subsec:supp_memorization}
\begin{figure}[ht]
    \includegraphics[width=\textwidth]{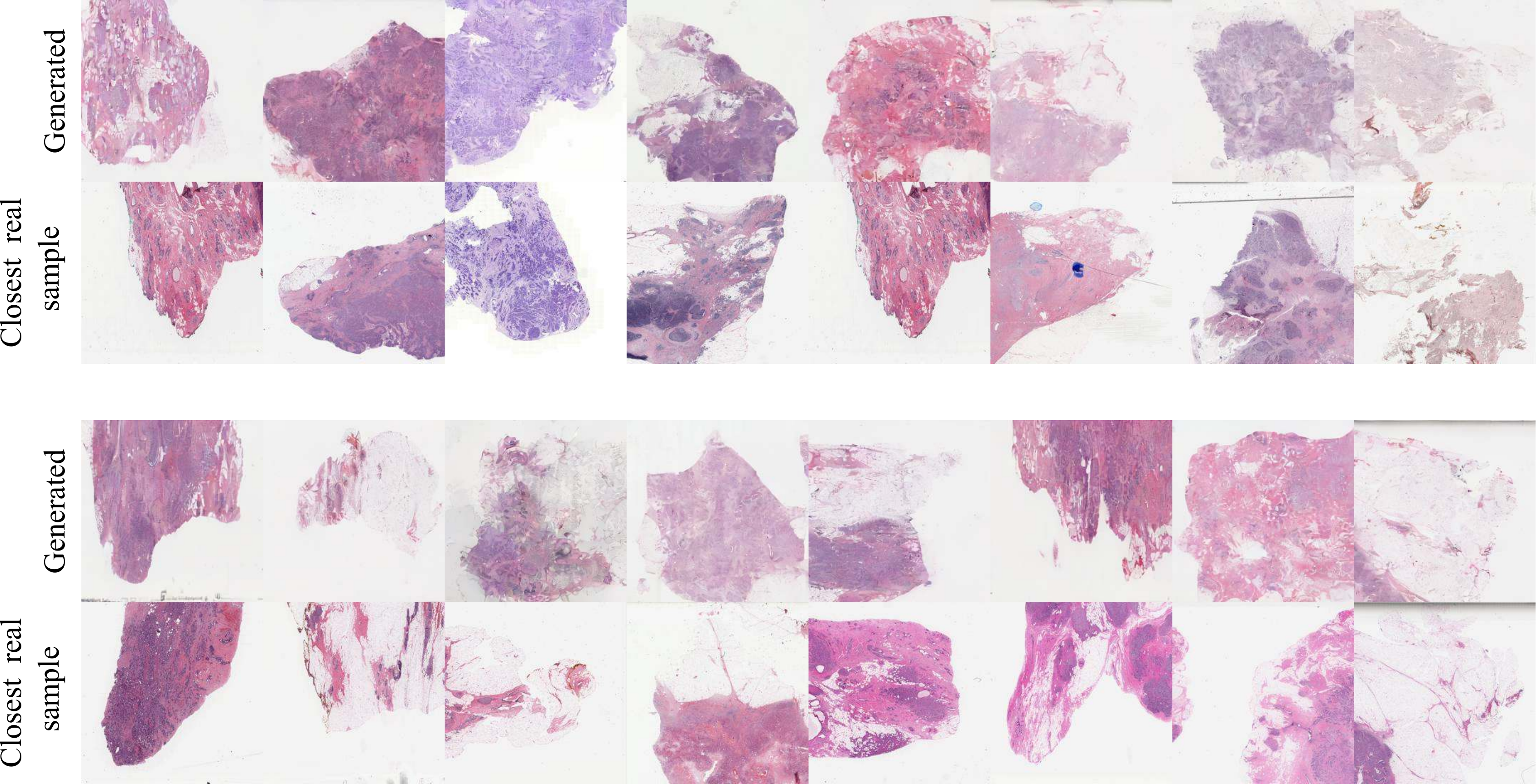}
    \centering
    \caption{We present $0.15625\times$ images generated from our model and their nearest neighbors in the training dataset. Although only trained on $\sim 2500$ images, our 400M parameter model did not memorize the training samples and successfully synthesized novel images at that magnification.}
    \label{fig:memorization}
\end{figure}

One of the arguments for training a single model for all scales is that we can learn to generate novel images even at scales with too few samples to learn from. To further demonstrate this, we use our histopathology diffusion model and sample conditions from the Conditioning Diffusion Model (CDM) to generate novel images at $0.15625\times$ magnification. At this scale, both our models have only seen $\sim 2500$ images and we would expect them to either generate low-quality samples or to have memorized the training data when using a 400M parameter model in training. Contrary to that, in Figure~\ref{fig:memorization}, we show that the generated images are realistic and different from the ones found in the training set. For each generated image, we identify its nearest neighbor in the training data using the patch-level UNI embeddings \cite{chen2024uni}, and show that they differ in shape and content. ZoomLDM can produce high-quality and unique samples for data-scarce magnifications, essentially avoiding memorization, by learning to synthesize images at all scales.

\subsection{Patches from all scales}
\label{subsec:supp_patches}
\begin{figure}[ht]
    \includegraphics[width=\textwidth]{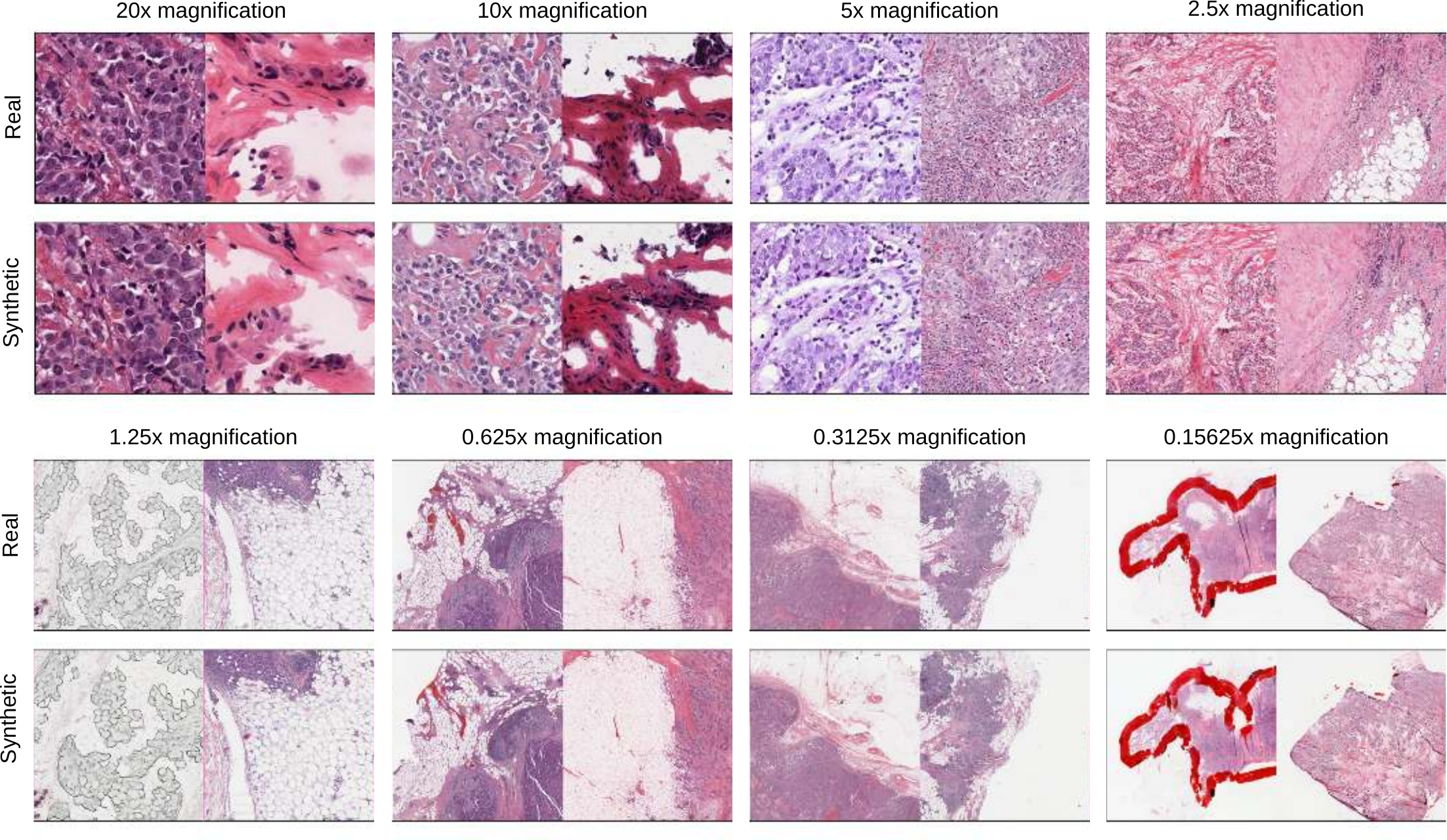}
    \centering
    \caption{Synthetic patches ($256 \times 256$ pixel) generated by ZoomLDM juxtaposed with the corresponding real images from TCGA-BRCA. Across all magnifications, ZoomLDM preserves the semantic features of the reference patches.}
    \label{fig:brca_all_mag}
\end{figure}

\begin{figure}[ht]
    \includegraphics[width=\textwidth]{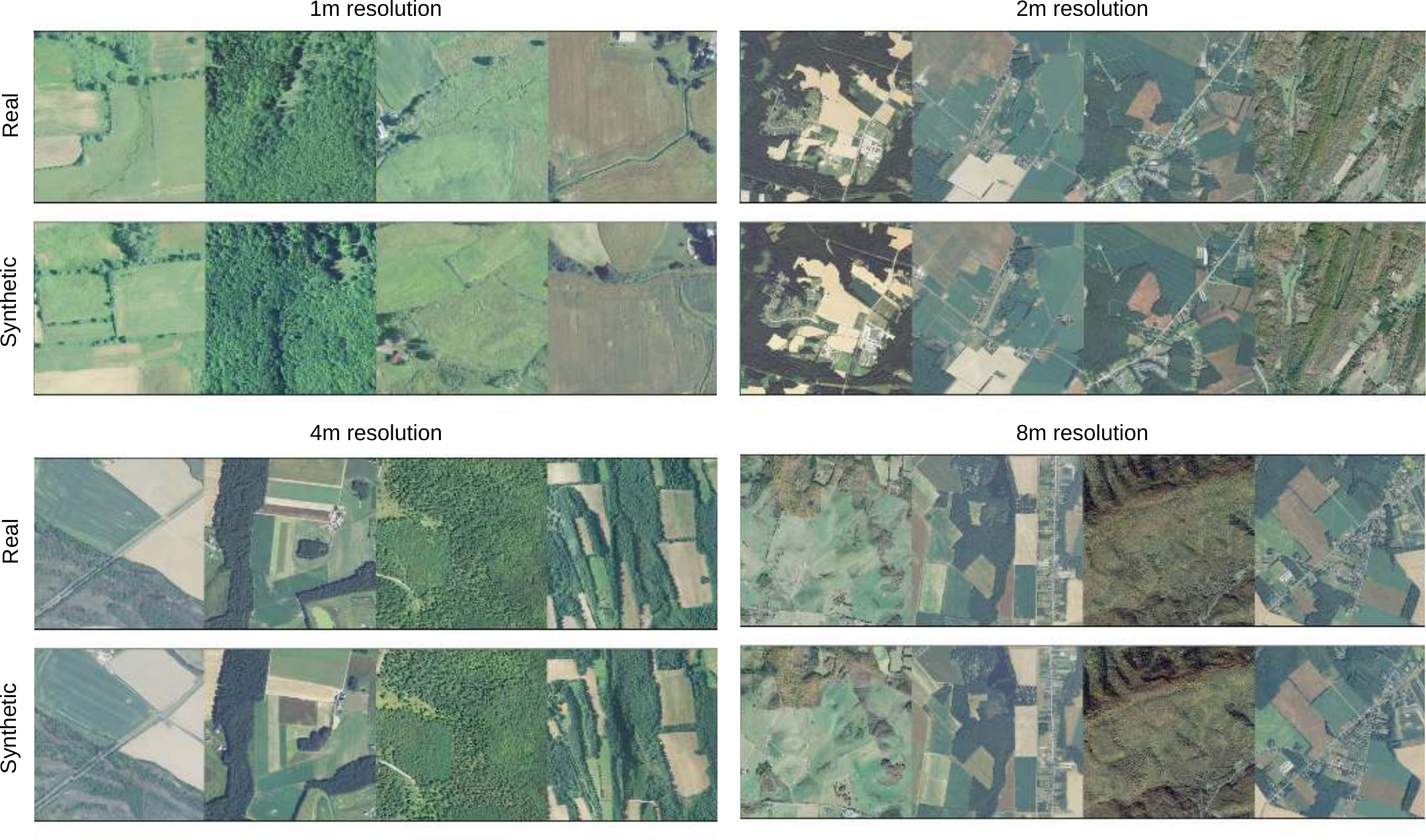}
    \centering
    \caption{Synthetic patches ($256 \times 256$ pixel) generated by ZoomLDM juxtaposed with the corresponding real images from NAIP}
    \label{fig:naip_all_mag}
\end{figure}

In Figures \ref{fig:brca_all_mag} and \ref{fig:naip_all_mag}, we showcase synthetic samples from ZoomLDM and the real images used to extract embeddings in histopathology and satellite. Samples from our model are realistic and preserve semantic features found in the reference patches. In data-scarce scenarios, such as $0.15625x$ magnification, achieving comparable image quality would be infeasible for a standalone model trained solely on that magnification (as indicated by the FIDs in Table 1 of the main text).

Interestingly, for magnifications below $5\times$ we find that the model can almost perfectly replicate the source image since the SSL embeddings used as conditioning contain enough information to reconstruct the patch at that scale perfectly. Although this may seem like a memorization issue, our experiments with the CDM in \ref{subsec:supp_memorization} show that our model has not just memorized the SSL embedding and image pairs. We believe that for these domains, this faithfulness to the conditions is advantageous as it can limit the hallucinations of the model, which are mostly unwanted in domains such as medical images.

\subsection{Large images}
\label{subsec:supp_large_images}
\begin{figure}[ht]
    \includegraphics[width=\textwidth]{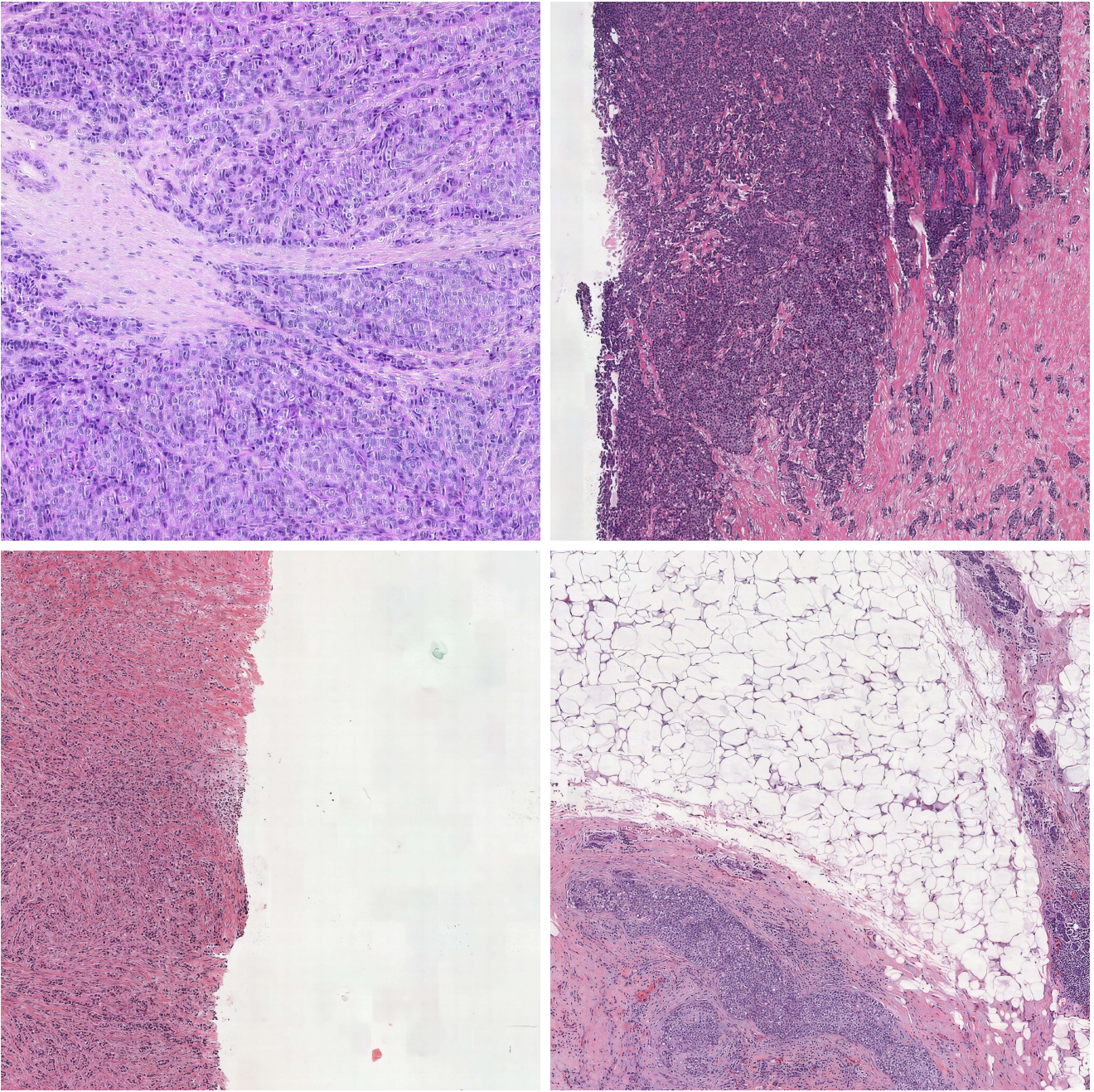}
    \centering
    \caption{We present $4096 \times 4096$ images generated from our histopathology model. Our results exhibit correct global structures in terms of the arrangement of cells and tissue while also maintaining high-resolution details. We point out two weaknesses: The local model fails to maintain coherency for structures where the lower-scale image does not provide guidance, such as the thin structures in the bottom-right image. In addition, for large uniform areas, such as the background in the bottom left image, the 'stitching' of the generated $20\times$ patches is visible with noticeable discontinuities along their edges.}
    \label{fig:brca_large}
\end{figure}

\begin{figure}[ht]
    \includegraphics[width=\textwidth]{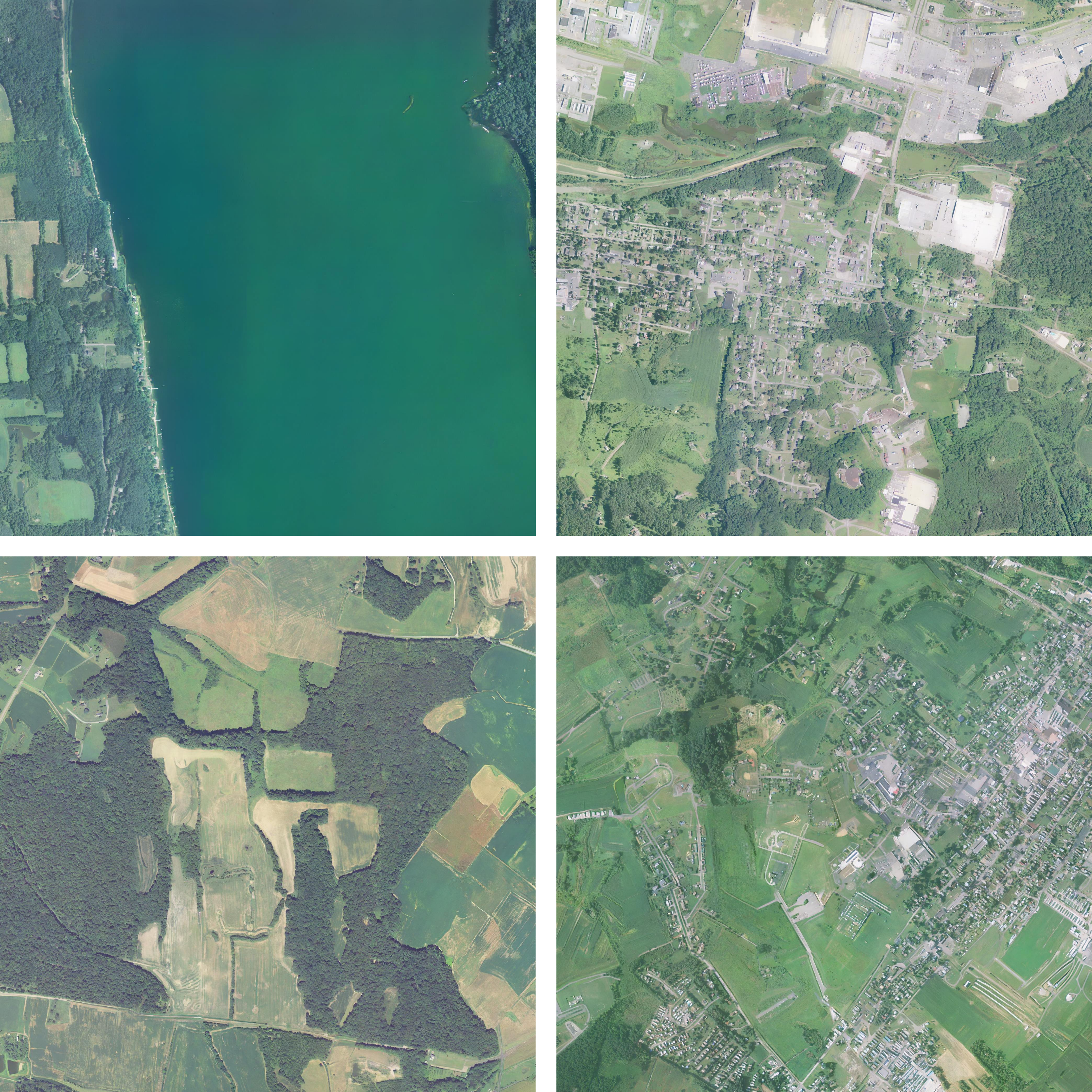}
    \centering
    \caption{We present $4096 \times 4096$ images generated from our satellite model. The results demonstrate images with reasonable global structures that also maintain high-resolution features. A similar weakness to the pathology images is visible, with slight discontinuities among the high-resolution patch borders.}
    \label{fig:naip_large}
\end{figure}

In Figures~\ref{fig:brca_large},\ref{fig:naip_large} we present $4096 \times 4096$ px images generated from our histopathology and satellite ZoomLDM model. Readers can find more examples on \href{https://histodiffusion.github.io/docs/projects/zoomldm}{histodiffusion.github.io/docs/projects/zoomldm}.

\subsection{Comparison to previous works}
\label{subsec:supp_prev_works}

\begin{figure}[ht]
    \includegraphics[width=\textwidth]{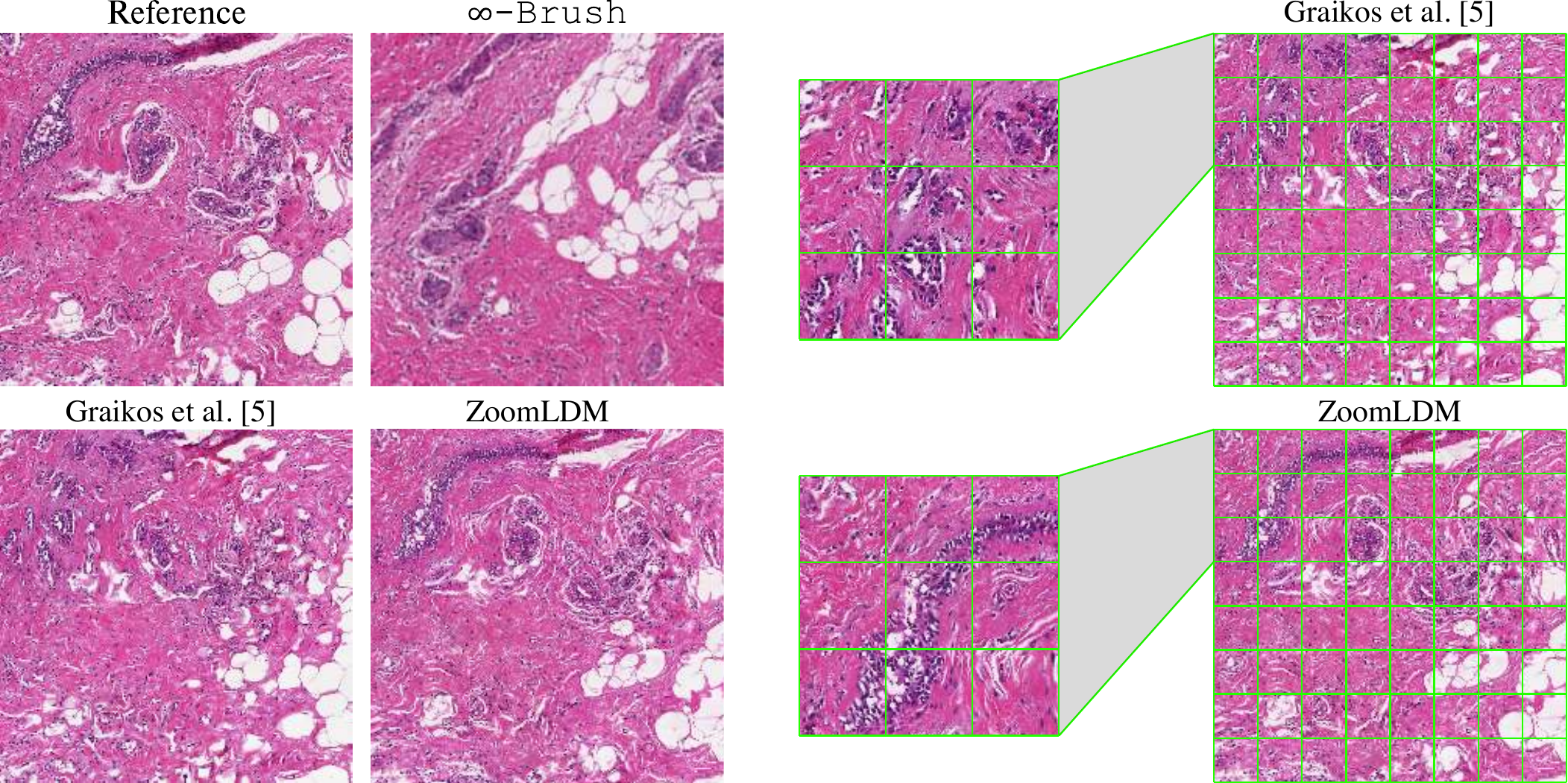}
    \centering
    \caption{We compare with two recent previous methods that also generated large histopathology images. In this example, we compare a $2048 \times 2048$ image from $\infty-\text{Brush}$ and \cite{graikos2024learned} to the same image generated from our model. We exceed both previous methods, with $\infty-\text{Brush}$ producing realistic global context but blurry details and \cite{graikos2024learned} completely failing to capture larger scale structures.}
    \label{fig:snake}
\end{figure}

\begin{figure}[ht]
    \includegraphics[width=\textwidth]{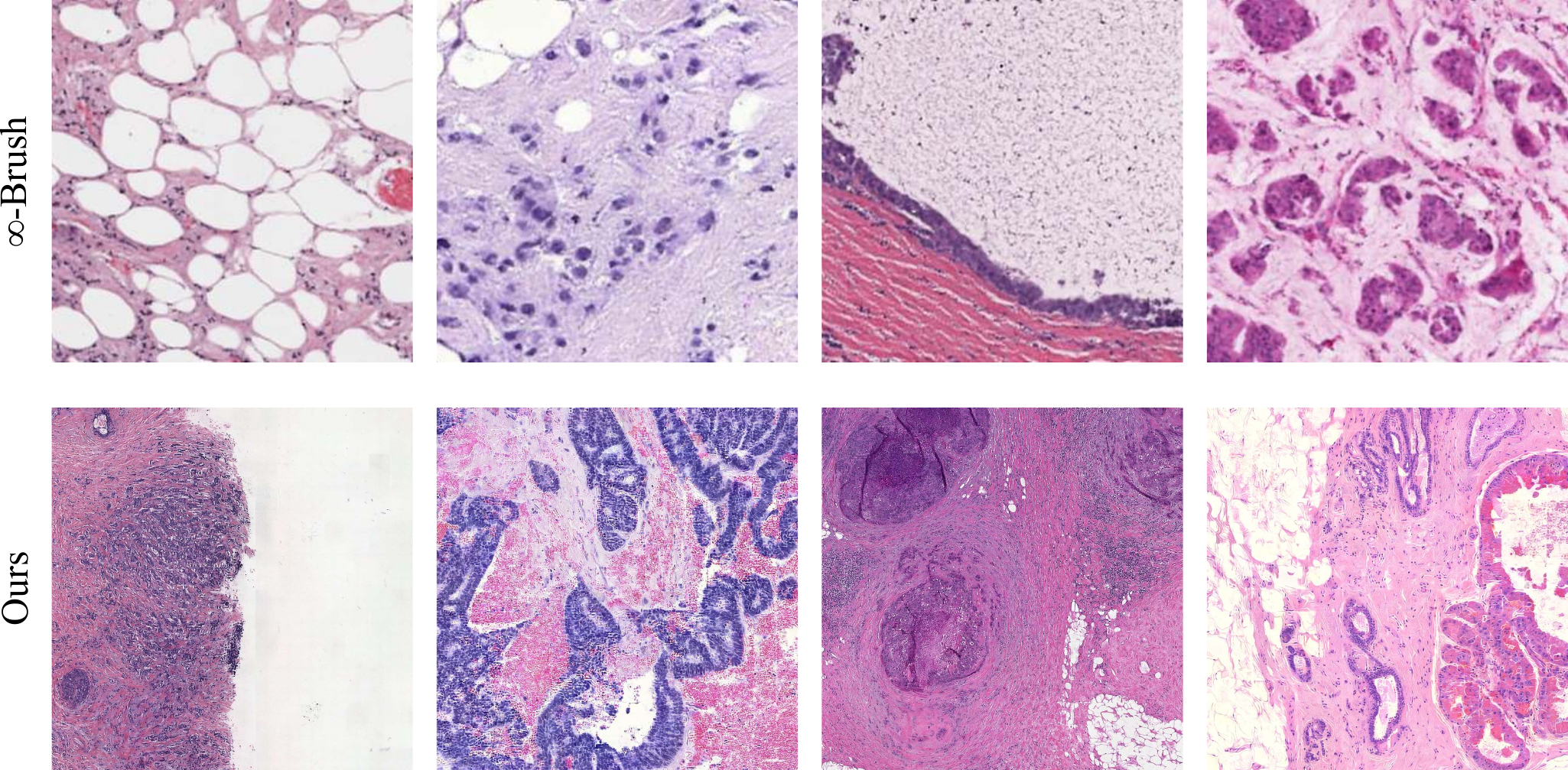}
    \centering
    \caption{Comparison between $\infty-\text{Brush}$ \cite{le2024inftybrushcontrollablelargeimage} and our method.}
    \label{fig:infty_brush_comparison}
\end{figure}

In Figure~\ref{fig:snake}, we compare our method and previous works on a single example image. We extract SSL embeddings from the 4k to replicate this image as closely as possible. We highlight our two main differences with previous methods. The method of $\infty-\text{Brush}$ \cite{le2024inftybrushcontrollablelargeimage} retains some global structures but fails to produce any high-resolution details in the image. On the other hand, the patch-based model of \cite{graikos2024learned} produces high-quality details but fails to capture large-scale structures that span more than a single patch. Our method solves both issues at the same time while maintaining a reasonable inference time, as discussed in the main text. We provide further comparisons to $\infty-Brush$ in Figure~\ref{fig:infty_brush_comparison}. Our generated images contain noticeably better detail.

\end{document}